\documentclass{article}
\usepackage[table]{xcolor}
\usepackage{iclr2027_conference,times}
\iclrfinalcopy

\usepackage{enumitem}

\usepackage[utf8]{inputenc} %
\usepackage[T1]{fontenc}    %
\usepackage{url}            %
\usepackage{booktabs}       %
\usepackage{amsfonts}       %
\usepackage{nicefrac}       %
\usepackage{microtype}      %
\usepackage{subcaption}
\definecolor{DarkGreen}{RGB}{1,100,32}
\definecolor{Purple2}{RGB}{100,0,100}

\usepackage{bm}
\usepackage{amsmath}
\usepackage{amssymb}
\usepackage{mathtools}
\usepackage{amsthm}

\usepackage{algorithm}
\usepackage{algpseudocode}
\usepackage{multirow}
\usepackage{wrapfig}
\usepackage{graphicx}
\usepackage{array}
\usepackage{needspace}
\newtheorem{theorem}{Theorem}

\definecolor{sidgray}{gray}{0.9}
\usepackage{xcolor}
\usepackage[most]{tcolorbox}
\usepackage[framemethod=tikz]{mdframed}
\usepackage{xparse}
\usepackage{tikz}

\newtcbox{\bracketeq}{
  nobeforeafter, enhanced, colback=gray!10, frame empty, math upper,
  boxsep=2pt, left=8pt, right=8pt, top=4pt, bottom=4pt, arc=4pt,
  tcbox raise base, %
  overlay={
    \def\thk{1.0pt}   %
    \def\stub{4pt}    %
    \draw[line width=\thk] (frame.north west) -- (frame.south west);
    \draw[line width=\thk] (frame.north east) -- (frame.south east);
    \draw[line width=\thk] (frame.north west) -- ([xshift=\stub]frame.north west);
    \draw[line width=\thk] (frame.south west) -- ([xshift=\stub]frame.south west);
    \draw[line width=\thk] (frame.north east) -- ([xshift=-\stub]frame.north east);
    \draw[line width=\thk] (frame.south east) -- ([xshift=-\stub]frame.south east);
  }
}

\newtcbox{\boxeq}{%
  enhanced,
  on line,
  nobeforeafter,
  math upper,          %
  frame empty,
  colback=gray!10,
  boxsep=2pt, left=4pt, right=4pt, top=4pt, bottom=4pt,
  arc=3pt,
  tcbox raise base,
  overlay={
    \def\r{3pt}         %
    \def\thk{1.0pt}     %
    \def\stub{4pt}      %
    \def\wipe{1pt}      %

    \draw[line width=\thk, rounded corners=\r]
      (frame.north west) rectangle (frame.south east);

    \fill[white]
      ([xshift=\stub,yshift=\wipe]frame.north west)
      rectangle
      ([xshift=-\stub,yshift=-\wipe]frame.north east);
    \fill[white]
      ([xshift=\stub,yshift=\wipe]frame.south west)
      rectangle
      ([xshift=-\stub,yshift=-\wipe]frame.south east);
  }
}
\usepackage{hyperref}       %

\definecolor{cafmblue}{HTML}{1D5B79}
\definecolor{cafmbluebg}{HTML}{EAF3F7}
\definecolor{priorblue}{HTML}{0858F8}
\definecolor{obsorange}{HTML}{FE5B0C}
\definecolor{agentpurple}{HTML}{8019BE}
\definecolor{dynteal}{HTML}{069170}
\newcommand{\priorTerm}[1]{\textcolor{priorblue}{#1}}
\newcommand{\obsTerm}[1]{\textcolor{obsorange}{#1}}
\newcommand{\agentTerm}[1]{\textcolor{agentpurple}{#1}}
\newcommand{\dynTerm}[1]{\textcolor{dynteal}{#1}}
\definecolor{reviewpurple}{HTML}{702C91}

\hypersetup{
  pdftitle={Training-Free Diffusion Planning with Analytical Local Scores},
  pdfauthor={Michael Yoo Fatemi, Jinhao Liang, Ferdinando Fioretto},
  colorlinks=true,
  linkcolor=cafmblue,
  citecolor=cafmblue,
  urlcolor=cafmblue
}

\title{Training-Free Diffusion Planning\\
with Analytical Local Scores}

\allowdisplaybreaks

\newcommand{\R}{\mathbb{R}}

\newcommand{\indicator}{\mathbb{I}}

\theoremstyle{definition}

\newcommand{\dif}{\mathop{}\!\mathrm{d}}

\author{Michael Yoo Fatemi\\
\normalfont University of Virginia\\
\href{mailto:gsk6me@virginia.edu}{\texttt{gsk6me@virginia.edu}}
\And
Jinhao Liang\\
\normalfont University of Virginia\\
\href{mailto:njs4nu@virginia.edu}{\texttt{njs4nu@virginia.edu}}
\And
Ferdinando Fioretto\\
\normalfont University of Virginia\\
\href{mailto:fioretto@virginia.edu}{\texttt{fioretto@virginia.edu}}}

\begin{document}

\maketitle
\fancyhead{}
\fancyfoot[L]{\footnotesize\textit{Fatemi, Liang, Fioretto - preprint}}
\fancyfoot[C]{}
\fancyfoot[R]{\thepage}
\renewcommand{\footrulewidth}{0.4pt}

\begin{abstract}
Motion planning requires trajectories that are smooth, goal-directed, and collision-free in environments with complex geometric constraints. Recent diffusion-based planners have shown that trajectory generation can be cast as iterative denoising, which has opened the doors to learning-based approaches that can handle multi-modal trajectory distributions and refine entire trajectories. 
However, such planners require training on large collections of feasible trajectories, rendering them map-specific, and difficult to deploy when high-quality demonstrations are unavailable. This paper introduces a training-free diffusion-based motion planner that replaces learned global trajectory scores with analytical local scores derived from obstacle, smoothness, velocity, and inter-agent feasibility terms. The proposed idea relies on a key observation: the trajectory score can be approximated from local interactions between neighboring waypoints and nearby constraints. This structure exploitation yields a decomposed denoising procedure that retains the optimization structure of classical optimization methods while inheriting the iterative refinement behavior of diffusion models. Experiments on a large collection of complex environments on multi-agent planning tasks show that the proposed analytical score produces smooth and feasible trajectories within very limited computational costs, for example in generating feasible paths for 300 agents in environments with 100 obstacles in under 6 seconds, while avoiding the data requirements of learned diffusion planners.
\end{abstract}

\section{Introduction}

Motion planning is a fundamental problem in robotics and autonomous systems, where the goal is to compute trajectories that move a system from an initial state to a target state while avoiding collisions with obstacles~\citep{lavalle_planning_2006}. In multi-agent settings, this problem becomes substantially more challenging: each robot must satisfy its own motion and safety constraints while also avoiding time-dependent collisions with other agents~\citep{sharon_conflict-based_2015,ma_searching_2018}. The spatio-temporal coupling and nonlinearity of these constraints make multi-agent path planning difficult to solve efficiently, especially as the number of agents and obstacles increases~\citep{liang_multi-agent_2024}.

Sampling-based planners provide strong completeness properties under suitable assumptions and have been widely used for high-dimensional planning problems. However, they often produce paths in a coarse configuration space that require additional smoothing and can become inefficient in narrow passages or crawded spaces~\citep{kuffner_rrt-connect_2000,lavalle_planning_2006}.
Optimization-based planners instead formulate motion planning as continuous trajectory optimization, making it possible to incorporate obstacle avoidance, smoothness, and kinematic constraints directly into the objective or feasible set~\citep{kalakrishnan_stomp_2011,mukadam_continuous-time_2018}. However, these methods are often sensitive to initialization and can become expensive in multi-agent settings because collision constraints couple many agents across time.
More recently, diffusion models have shown promising results for motion planning by casting trajectory generation as an iterative denoising process that refines noise into trajectories~\citep{liang_discrete-guided_2025,shaoul_multi-robot_2025}. This perspective is attractive because diffusion models refine entire trajectories rather than committing to one waypoint at a time, and can represent multimodal distributions over possible plans~\citep{carvalho_motion_2023}. The key challenge, however, is that existing diffusion planners require learning a neural score model over high-dimensional trajectory distributions. This makes them expensive to train, dependent on large collections of feasible demonstrations, and often map- or distribution-specific. Moreover, learned diffusion planners may still fail to enforce safety constraints, a critical limitation for robotic planning~\citep{liang_multi-agent_2024}. Projection-based refinement can improve constraint satisfaction, but it introduces additional computational cost~\citep{liang_discrete-guided_2025,liang_multi-agent_2024}.

\begin{figure}[!t]
    \centering
    \includegraphics[width=\linewidth]{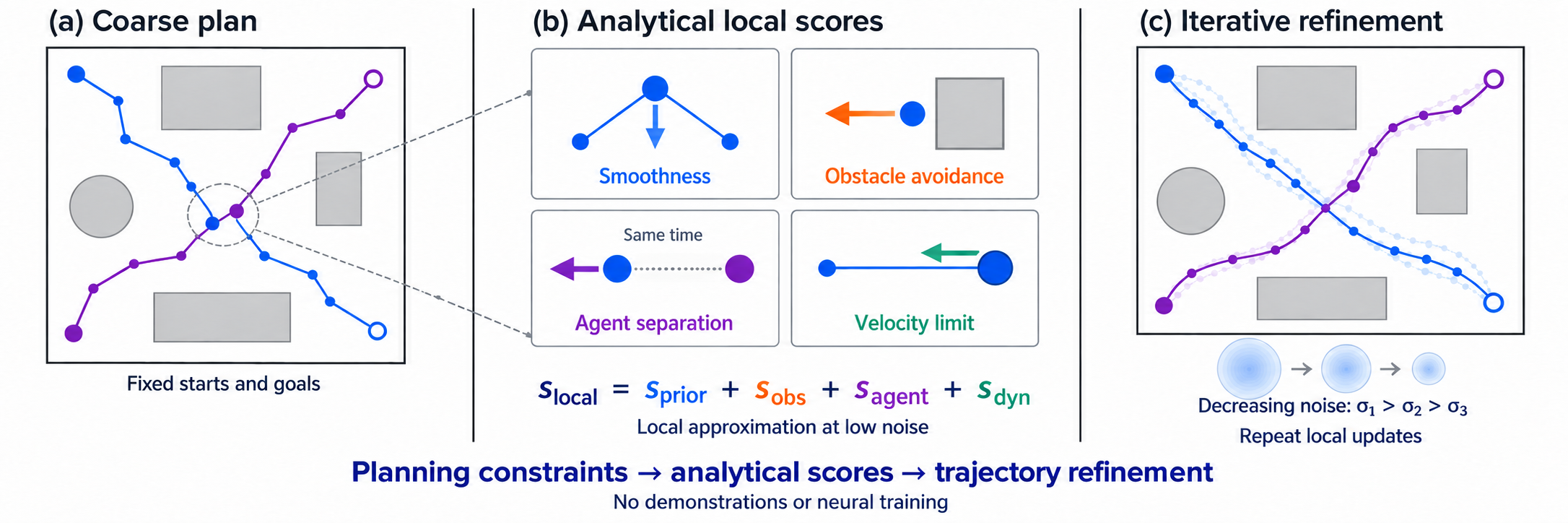}
    \caption{\textbf{Training-free diffusion planning.}
    (a) Coarse trajectories connect fixed starts and goals.
    (b) Analytical local scores combine (i) smoothness, (ii)obstacle avoidance, (iii) inter-agent separation, and (iv) velocity limits.
    (c) Repeated updates under decreasing noise refine the trajectories without neural training.
    [Note: the refined paths cross spatially but pass the intersection at different times].}
    \label{fig:overview}
\end{figure}

To address these challenges, this paper develops a training-free diffusion-based planner with fully analytical local scores. The key idea is based on the observation that trajectory refinement has a strong locality of influence~\citep{pan2012collision}. Although the exact score of a full trajectory distribution can be globally coupled, the update of an individual waypoint is dominated by local temporal and spatial interactions: neighboring waypoints determine smoothness and velocity consistency, nearby obstacles determine collision avoidance, and nearby agents at the same time determine inter-robot separation~\citep{mukadam2018continuous,zucker_chomp_2013}, as illustrated in Figure~\ref{fig:overview}(b).
Motivated by this structure, we introduce a decomposed local score formulation that constructs an analytical score surrogate directly from the planning constraints, rather than learning a neural approximation to the score of a high-dimensional distribution over complete trajectories. The resulting planner starts from a coarse trajectory initialization with fixed endpoints (Figure~\ref{fig:overview}(a)) and repeatedly applies analytical denoising updates under a decreasing noise schedule (Figure~\ref{fig:overview}(c)). Because each update depends only on local trajectory neighborhoods, nearby obstacles, and nearby agents, the method avoids dataset collection and neural-network training while remaining scalable to large multi-agent scenes.
\\[4pt]
\textbf{Contributions.}
The paper makes the following contributions:
\textbf{(1)} It introduces a training-free diffusion planning framework that replaces learned global trajectory scores with analytical score surrogates derived from the structure of the motion planning problem.
\textbf{(2)} It develops a decomposed local score formulation in which each waypoint is refined using smoothness, obstacle-avoidance, velocity-feasibility, and inter-agent collision-avoidance terms.
\textbf{(3)} It demonstrates across multi-agent path planning benchmarks that the proposed planner achieves high feasibility and computational efficiency, obtaining high success on standard test configurations and scaling to scenes with $300$ agents and $100$ obstacles in just a few seconds.

\section{Related Work}
\vspace{-4pt}

\textbf{Multi-agent path planning} has a long history in robotics, with early methods largely built around sampling-based and optimization-based approaches. \emph{Sampling-based planners}, approximate the feasible configuration space by drawing samples and connecting them through local steering procedures~\citep{kavraki_probabilistic_1996,kuffner_rrt-connect_2000,lavalle_planning_2006}. These can scale to high-dimensional spaces better than exhaustive search methods, however, they often produce piecewise or fragmented paths that require post-processing, and their performance can degrade in narrow passages or settings where continuous kinodynamic constraints must be enforced along the entire trajectory. 
A complementary line of work formulates motion planning as \emph{trajectory optimization} problems%
~\citep{6385823,9197162,6225009}. This gives these methods a direct mechanism for incorporating feasibility requirements, but the resulting optimization problems become increasingly difficult in multi-agent settings, where collision constraints couple agents across both space and time. To improve computational tractability, several optimization-based planners replace hard feasibility constraints with differentiable costs or penalty terms~\citep{ratliff_chomp_2009,kalakrishnan_stomp_2011,mukadam_continuous-time_2018}, but they are often sensitive to initialization and can struggle in environments with many obstacles or agents. In contrast, our method formulates planning as an iterative denoising process driven by analytical local scores, avoiding dense state-space exploration while also avoiding repeated global trajectory optimization.
\\[4pt]
\textbf{Score-based Generative Models for Motion Planning}.
Recent advancements in generative models offer have opened new avenues for solving motion planning problems by treating feasible trajectories as samples from a distribution and generating them through iterative denoising~\citep{song2020score}. Diffusion planners are attractive because they refine whole trajectories at once and can represent multimodal plans in dense environments~\citep{carvalho_motion_2023,shaoul_multi-robot_2025,luo2024potential}. However, they typically require learning a neural score over high-dimensional trajectory distributions, which depends on large collections of feasible demonstrations and does not guarantee constraint satisfaction. To improve feasibility, recent work has added gradient-based guidance during sampling, control barrier and Lyapunov functions, or projection-based refinement after denoising~\citep{carvalho_motion_2023,mizuta_cobl_2024,shaoul_multi-robot_2025,luo2024potential,liang_discrete-guided_2025,liang_multi-agent_2024,Fioretto:NeurIPS24,fioretto:26h}. These  Both retain learned generative models; TFDP constructs its refinement field directly from smoothness and feasibility factors.

The method proposed in this work, instead, replaces the learned global score with analytical local score components derived directly from smoothness, velocity, obstacle-avoidance, and collision constraints. This yields a training-free planner that preserves the iterative refinement behavior of diffusion models while making denoising updates fast and tied to the geometry of the environment.

\section{Preliminaries}

\vspace{-4pt}
\textbf{Multi-agent path planning.} 
Multi-agent path planning involves computing collision-free trajectories for multiple agents navigating a shared environment from designated start positions to goal states. 
Let $\mathcal{A}=\{a_1,\dots,a_{N_a}\}$ be a set of $N_a$ robots that move
in a bounded, planar workspace $\mathcal{W}\subset\mathbb{R}^2$ containing
a set of static obstacles $\mathcal{O}=\{o_1,\dots,o_{N_o}\}$.
The \emph{free configuration space} is defined as the set of all points in the workspace that are not occupied by obstacles:
\(
   \mathcal{C}_{\mathrm f}
   := \mathcal{W}\setminus
      \bigl(\bigcup_{o\in\mathcal{O}} o\bigr).
\) 
Each robot $a_i$ is modeled as a disk of radius $r_i>0$ and is characterized at discrete time steps $h\in\{0,\ldots,H\}$ by a 2-D position $\pi_i^{h} = (x_i^{h},y_i^{h})\in\mathcal{C}_{\mathrm f}.$ A \emph{trajectory} for robot $a_i$ is the sequence $\bm{\pi}_i := (\pi_i^{0},\pi_i^{1},\dots,\pi_i^{H}).$

The goal of multi-agent path planning is to compute a set of trajectories $\boldsymbol{\Pi} = \{\boldsymbol{\pi}_1, \boldsymbol{\pi}_2, \ldots, \boldsymbol{\pi}_{N_a}\}$ such that each robot starts at its designated start position and reaches its target position while avoiding collisions with obstacles and other agents.
For every time index $h$ we require
\begin{subequations}
\begin{align}
  \pi_i^{h} &\in \mathcal{C}_{\mathrm f},
  &\forall i,
  \label{eq:obst-avoid}\\[2pt]
  \|\pi_i^{h}-\pi_j^{h}\|_2 &\ge r_i+r_j,
  &\forall i<j,          \label{eq:robot-avoid}\\[2pt]
  \|\pi_i^{h+1}-\pi_i^{h}\|_2 &\le v_{\max}\,\Delta t,
  &\forall i,h<H,        \label{eq:dyn}
\end{align}
\end{subequations}
where Eq.~~\eqref{eq:obst-avoid} enforces obstacle avoidance,
Eq.~\eqref{eq:robot-avoid} inter-robot separation, and Eq.~\eqref{eq:dyn} a
first-order kinematic bound with time step~$\Delta t$ and speed
limit~$v_{\max}$.

\textbf{Diffusion models for motion planning.}
Score-based diffusion models define a forward noising process and a reverse
sampling process~\citep{song2020score}. In the forward process, samples from a
data distribution are gradually perturbed with Gaussian noise. Let $\pi$ denote
a trajectory, and let $p_0(\pi \mid \mathcal{O},s,g)$ denote the distribution
of feasible trajectories conditioned on the obstacle set $\mathcal{O}$, start
state $s$, and goal state $g$. At noise level $\sigma$, the forward process
induces a noise-smoothed distribution $p_\sigma$ by
$\pi_\sigma = \pi + \sigma \epsilon$, with $\epsilon \sim \mathcal{N}(0,I)$.
The reverse process then generates samples by moving noisy trajectories back
toward the feasible trajectory distribution. This reverse process is driven by
the score function $\nabla_\pi \log p_\sigma(\pi \mid \mathcal{O},s,g)$, 
which points in the direction of increasing probability under the noise-smoothed distribution. 
In learned diffusion planners, feasible trajectories are treated as data samples
and used to train a neural network $s_\theta$ to approximate this score
function~\citep{carvalho_motion_2023,shaoul_multi-robot_2025,
liang_multi-agent_2024}:
\begin{equation}
    \label{eq:score_approx}
    s_\theta(\pi;\mathcal{O},s,g). \approx \nabla_\pi \log p_\sigma(\pi \mid \mathcal{O},s,g)
\end{equation}
At sampling time, the learned score is used in Langevin dynamics to iteratively
refine an initial noisy trajectory, for example through updates of the form
\[
    \pi \leftarrow \pi
    + \epsilon  s_\theta(\pi;\mathcal{O},s,g)
    + \sqrt{2 \epsilon \tau }\,\xi,
    \qquad \xi\sim\mathcal{N}(0,I),
\]
with a decreasing noise schedule, step size $\epsilon$ (typically chosen proportional to $\sigma^2$), and optional temperature parameter $\tau$ \citep{xu_global_2018}. Thus, the learned score determines the
direction in which the full trajectory is refined at each denoising step.

Crucially, the score network $s_\theta$ is a function of the \emph{entire noisy trajectory}, since feasibility depends on global trajectory structure, obstacle geometry, and interactions among agents across time. Prior diffusion planners often add guidance terms during sampling to penalize obstacle penetration, inter-agent collisions, or trajectory roughness~\citep{carvalho_motion_2023,liang_multi-agent_2024,liang_discrete-guided_2025}. However, the underlying model still requires learning a global score over high-dimensional trajectory distributions, which is expensive and depends on access to feasible training trajectories.
Next, the paper introduces a method that avoids learning such a global score model altogether.

\section{Training-Free Diffusion-based Planning (TFDP)}
\label{sec:tfdp}

This section presents \emph{Training-Free Diffusion-based Planning} (TFDP),
a method that performs iterative denoising without a learned score model.
The key idea is to replace the unknown global score
$\nabla_\pi \log p_\sigma(\pi \mid \mathcal{O}, s, g)$ of the data distribution
with the score of an explicitly \emph{constructed} surrogate target, whose factors are dictated by the locality of the planning problem itself. 

\subsection{A Surrogate Target Distribution over Trajectories}
\label{sec:tfdp:target}

Rather than learning to approximate the data distribution of feasible trajectories, TFDP constructs a surrogate that exploits the spatio-temporal structure of the path planning problem. The surrogate is a product of four factors, colored in Figure~\ref{fig:overview}(b), that encode smoothness, obstacle avoidance, inter-agent separation, and kinematic feasibility:
\begin{equation}
  p(\pi) \;\propto\; \priorTerm{\phi_{\mathrm{prior}}(\pi)}\,
                     \obsTerm{\chi_{\mathrm{obs}}(\pi)}\,
                     \agentTerm{\chi_{\mathrm{agent}}(\pi)}\,
                     \dynTerm{\chi_{\mathrm{dyn}}(\pi)},
  \label{eq:surrogate}
\end{equation}
Specifically, the Gaussian factor
\begin{equation}
    \label{eq:smoothness_prior}
  \priorTerm{\phi_{\mathrm{prior}}(\pi)}
  \;=\; \exp\!\left(
        -\tfrac{\lambda_{\mathrm{sm}}}{2}
        \sum_{i}\sum_{h=1}^{H-1}
        \bigl\|\pi^{h+1}_i - 2\pi^h_i + \pi^{h-1}_i\bigr\|^2
        \right),
\end{equation}
penalizes sharp bends through a Gaussian finite-difference smoothness prior, derived from locality of motion and from the desire of producing smooth trajectories.  The other three indicator factors are
\begin{subequations}
\begin{align}
  \obsTerm{\chi_{\mathrm{obs}}(\pi)}
    &= \prod_i \prod_{h=1}^{H}
       \indicator\!\left[\,\pi^h_i \in \mathcal{C}_f\,\right], \label{eq:indicator_obs} \\
  \agentTerm{\chi_{\mathrm{agent}}(\pi)}
    &= \prod_{i<j} \prod_{h=1}^{H}
       \indicator\!\left[\,\bigl\|\pi^h_i - \pi^h_j\bigr\| \geq r_i + r_j\,\right], \label{eq:indicator_agent} \\
  \dynTerm{\chi_{\mathrm{dyn}}(\pi)}
    &= \prod_i \prod_{h=1}^{H-1}
       \indicator\!\left[\,\bigl\|\pi^{h+1}_i - \pi^h_i\bigr\| \leq v_{\max}\Delta t\,\right] \label{eq:indicator_dyn}.
\end{align}
\end{subequations}
The three indicator factors are ordered to match the feasibility constraints: Eq.~\eqref{eq:indicator_obs} corresponds to obstacle avoidance in Eq.~\eqref{eq:obst-avoid}, Eq.~\eqref{eq:indicator_agent} corresponds to inter-robot separation in Eq.~\eqref{eq:robot-avoid}, and Eq.~\eqref{eq:indicator_dyn} corresponds to the kinematic bound in Eq.~\eqref{eq:dyn}.
The surrogate~\eqref{eq:surrogate} deliberately discards information about
which feasible trajectories are \emph{likely} under demonstration data,
retaining only which trajectories are \emph{feasible and smooth}. 
This is the price paid for being training-free, but it is also a virtue: the surrogate is entirely determined by the problem geometry and the smoothness weight, so it can be evaluated without any learned component. The next step is to show how to compute its score in closed form.

Then, for each waypoint $\pi_i^h$ of agent $a$, we approximate the full score at noise level $\sigma$ by
\begin{equation}
\begin{aligned}
\nabla_{\pi_i^h} \log p_\sigma(\pi \mid \mathcal{O})
\;\approx\;{}&
\priorTerm{s_{\mathrm{prior}}(\pi_i^h;\, \pi_i^{h-1}, \pi_i^{h+1})}
+ \obsTerm{s_{\mathrm{obs}}(\pi_i^h;\, \mathcal{O})} \\
&+ \agentTerm{s_{\mathrm{agent}}(\pi_i^h;\, \pi_{j\neq i}^{h})}
+ \dynTerm{s_{\mathrm{dyn}}(\pi_i^h;\, \pi_i^{h-1}, \pi_i^{h+1})}.
\end{aligned}
\label{eq:local-score}
\end{equation}
The four terms define the local score ($s_{\mathrm{local}}$ in Figure~\ref{fig:overview}(b)) combining smoothness, agent-obstacle collision avoidance, agent-agent collision avoidance, and dynamic feasibility (i.e., velocity constraints). The approximation removes long-range score dependencies while preserving the terms that dominate local feasibility. Note this decomposition also makes the update interpretable as each waypoint is moved according to the direction described by each multi-agent path planning constraint. 
We describe the explicit form of the score components in Section \ref{sec:analytical_score_terms}.

The local score decomposition approximates the noisy trajectory score and does not imply exact conditional independence between distant waypoints. For Gaussian trajectory priors with local interactions, zeros in the precision matrix encode conditional independencies, and adding isotropic diffusion noise changes the precision structure. At small noise levels, indirect dependencies between distant variables are higher-order effects relative to the direct neighbor interactions. This motivates using short score windows in low-noise denoising regimes and considering larger windows at higher noise levels. 
We formalize this intuition next, by showing that the conditional independence structure of a multivariate Gaussian is approximately preserved under diffusion noise. This result is of independent interest and may be useful in other applications of diffusion models.
\begin{theorem}%
\label{thm:Independence}
Let $x \sim \mathcal{N}(0, \Sigma)$ be a multivariate normal distribution with covariance $\Sigma \in \mathbb{R}^{n \times n}$ and precision matrix $\Omega := \Sigma^{-1}$. Let $\{x_i\}_{i=1}^n$ denote the components, and consider the associated Markov random field (MRF) where an edge $(i,j)$ is present iff $\Omega_{ij} \neq 0$.
\\[4pt]
Consider a diffusion process applied to $x$ with noise level $t$, and let $I(x_i; x_j \mid x_{-\{i, j\}})$ denote the mutual information between components $x_i$ and $x_j$. Then the rate of increase of mutual information at small $t$ is given by
\begin{equation}
\frac{\mathrm{d}}{\mathrm{d}t} I(x_i; x_j \mid x_{-\{i, j\}})
= t \frac{\left|\sum_{k} \Omega_{ik} \Omega_{kj} \right|^2}{\Omega_{ii} \Omega_{jj}}.
\end{equation}
Consequently, the leading-order contribution to mutual information between $x_i$ and $x_j$ is governed by length-2 paths in the MRF. Variables separated by longer paths incur dependence only at higher orders in $t$, and are therefore negligible for sufficiently small $t$.
\end{theorem}
The detailed proof is shown in Appendix~\ref{appendix:conditional_information_flow_derivation}. Theorem~\ref{thm:Independence} shows that the local independence structures are approximately preserved. This is important for our proposed method because it motivates computing score updates from local trajectory neighborhoods at low noise levels.

\subsection{Analytical Score Terms}
\label{sec:analytical_score_terms}
\vspace{-4pt}
Now, we provide explicit expressions for $s_{\priorTerm{\mathrm{prior}},\obsTerm{\mathrm{obs}},\agentTerm{\mathrm{agent}},\dynTerm{\mathrm{dyn}}}$. 
We first introduce the constraint set: 
\begin{subequations}
\begin{align}
\mathcal{C}_{\mathrm{obs}}(\pi_i^h; \mathcal{O}) := \ & \mathcal{C}_{\mathrm f}, \\
\mathcal{C}_{\mathrm{agent}}(\pi_i^h; \pi_{i \neq j}^h) := \ & \left\{
y\in\mathcal W:
\|y-\pi_j^h\|_2 \ge r_i+r_j,\ \forall j\neq i
\right\}, \label{eq:local-C-agent} \\
\mathcal{C}_{\mathrm{dyn}}(\pi_i^h; \pi_i^{h + 1}, \pi_i^{h - 1})  := \ & \left\{
y\in\mathcal W:
\|y-\pi_i^{h-1}\|_2 \le v_{\max}\Delta t,\ 
\|\pi_i^{h+1}-y\|_2 \le v_{\max}\Delta t
\right\}, \label{eq:local-C-dyn}
\end{align}
\end{subequations}
where $y\in\mathcal W$ denotes a candidate position for the waypoint $\pi_i^h$. Using the smoothness factor defined in Equation \eqref{eq:smoothness_prior}, $\priorTerm{s_{\mathrm{prior}}} = \nabla \log \priorTerm{\phi_{\mathrm{prior}}(\pi)}$. For the constraint components, let:
\begin{subequations}
\begin{align}
\obsTerm{\chi_{\mathrm{obs}}^\sigma(\pi_i^h; \mathcal{O})}
&= K_\sigma * \mathcal{C}_{\mathrm{obs}}(\pi_i^h; \mathcal{O}) \\
\agentTerm{\chi_{\mathrm{agent}}^\sigma(\pi_i^h; \pi_{i \neq j}^h)}
&= K_\sigma * \mathcal{C}_{\mathrm{agent}}(\pi_i^h; \pi_{i \neq j}^h)\\
\dynTerm{\chi_{\mathrm{dyn}}^\sigma(\pi_i^h; \pi_i^{h + 1}, \pi_i^{h - 1})}
&= K_\sigma * \mathcal{C}_{\mathrm{dyn}}(\pi_i^h; \pi_i^{h + 1}, \pi_i^{h - 1}) 
\end{align}
\end{subequations}
be the constraint indicators convolved with Gaussians; i.e., a closed-form solution to the forward diffusion process of a uniform distribution over the corresponding feasible set, up to a constant. Then we use $s_{\obsTerm{\mathrm{obs}},\agentTerm{\mathrm{agent}},\dynTerm{\mathrm{dyn}}}(\cdot) = \nabla \log \chi_{\obsTerm{\mathrm{obs}},\agentTerm{\mathrm{agent}},\dynTerm{\mathrm{dyn}}}^{\sigma}(\cdot)$, which pull the sample towards the feasible region with increasing strength as $\sigma \to 0$. Unlike guidance terms, the constraint terms $\nabla \log \obsTerm{\chi^\sigma_{\mathrm{obs}}}$ are explicitly tied to the probability that a noisy waypoint is feasible. Computing these terms directly requires a convolution integral. For a feasible set $\mathcal{C}_{\mathrm{free}}$ of any arbitrary constraint,
\begin{equation}
\chi^\sigma_{\mathrm{constraint}}(\pi_i^h) = \int_{\mathcal{C}_\mathrm{free}} \mathcal{N}(\pi_i^h; x, \sigma^2 I) \dif x
\end{equation}
We provide closed-form solutions for circular and rectangular feasible sets in Appendix \ref{appendix:analytical_score_derivation}, \ref{appendix:rectangular_box_score_function_derivation}, and \ref{appendix:rectangular_infeasible_region_score_function_derivation}, respectively, however, closed-form solutions are not available for general, non-convex sets, for which we propose a low-noise approximation that avoids computing the convolution integral directly.
\begin{theorem}[Low-noise score approximation and projection]
Let $p_\sigma(x)$ denote a Gaussian-smoothed density with noise scale $\sigma > 0$, and let $\Pi(x)$ denote the Euclidean projection of $x$ onto a feasible set $\mathcal{C}_{\mathrm{f}}$. Then, in the limit as $\sigma \to 0$, the score function satisfies the first-order approximation
\begin{equation}
\sigma^2 \nabla \log p_{\sigma}(x) \approx \bigl(\Pi(x) - x\bigr).
\end{equation}
Thus, the score is proportional (up to a factor of $\sigma^2$) to the negative gradient of the squared distance to the feasible set. In particular, letting $d(x, \mathcal{C}_{\mathrm{f}})$ denote the Euclidean distance to $\mathcal{C}_{\mathrm{f}}$, we have
\begin{equation}
\nabla \left( \tfrac{1}{2} d(x, \mathcal{C}_{\mathrm{f}})^2 \right)
= x - \Pi(x),
\end{equation}
which implies that the score direction aligns with the gradient of $\frac{1}{2} d(x, \mathcal{C}_{\mathrm{f}})^2$ up to scaling.
\end{theorem}

\begin{figure}[t]
    \centering
    \includegraphics[width=0.8\linewidth]{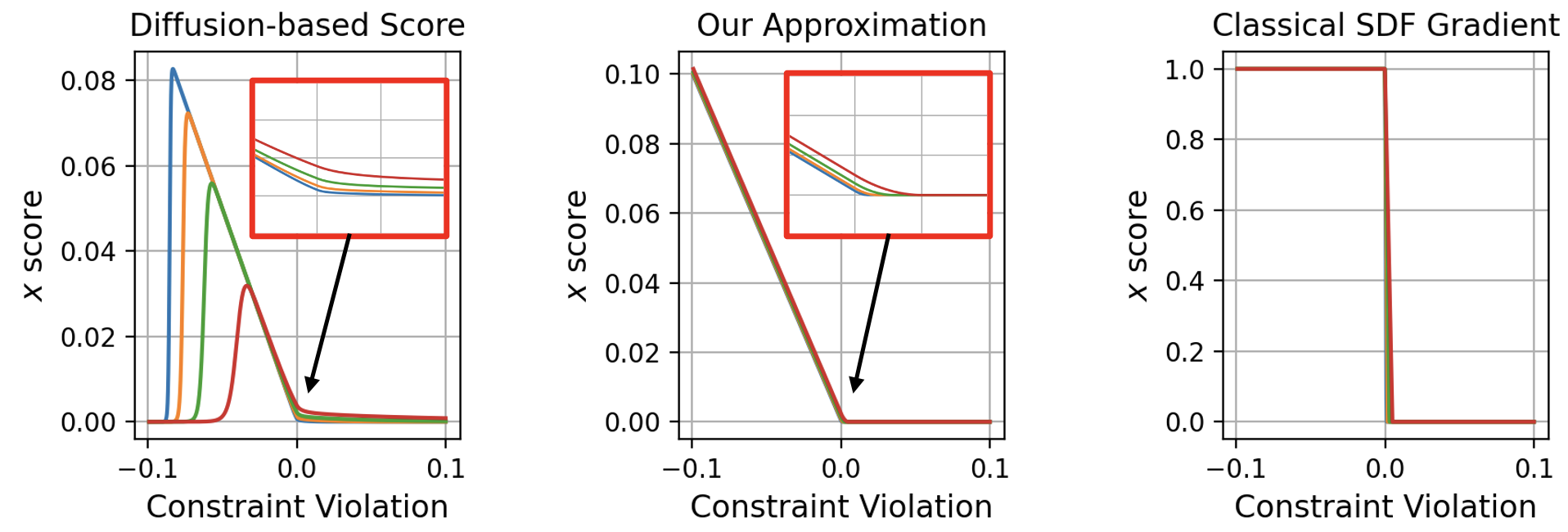}
    \caption{Effective guidance from score candidates}
    \label{fig:sdf_vs_diffusion_score}
    \vspace{-12pt}
\end{figure}
See Appendix~\ref{appendix:projection_at_low_sigma} for the proof. 
Note that, due to the smoothing effect of the Gaussian convolution, the score function is nonzero even for points inside the feasible region, for small distances.
Therefore, we also include a quadratically-scaled gradient such that the vertex of the parabola is at distance $\sigma$ from the obstacle boundary. This motivates a general-purpose constraint satisfaction score function, constraining $\pi_i^h \in \mathcal{C} \to \pi_i^h \in \mathcal{C}_{\mathrm{f}} \, \forall \, i, h$:
\begin{equation}
\psi^{\sigma}(\pi_i^h, \mathcal{C}_f)
= \nabla d(\pi_i^h, \mathcal{C}_f) \cdot
\begin{cases}
\, d(\pi_i^h, \mathcal{C}_f) + \dfrac{\sigma}{2}
   & \text{if } d(\pi_i^h, \mathcal{C}_f) > 0, \\[8pt]
\, \nicefrac{1}{2\sigma}\!\left( d(\pi_i^h, \mathcal{C}_f) + \sigma \right)^{2}
   & \text{if } d(\pi_i^h, \mathcal{C}_f) \in (-\sigma,\, 0], \\[8pt]
\, 0 & \text{otherwise.}
\end{cases}
\label{eq:psi}
\end{equation}
\begin{wrapfigure}[12]{r}{0.4\linewidth}
\vspace{-1.0\baselineskip}
\footnotesize
\begin{minipage}{\linewidth}
\setlength{\intextsep}{0pt}
\begin{algorithm}[H]
    \small
    \caption{TFAD planner}\label{alg:analytical-diffusion-planner}
    \begin{algorithmic}[1]
    \State \textbf{Input:} start $s$, goal $g$, schedule $\{\sigma_k\}_{k=1}^K$, step size $\eta$, temp.~$\tau$
    \State Initialize $\pi$ with a coarse roadmap planner
    \For{$k=1,\ldots,K$}

        \State Sample $\xi_k \sim \mathcal{N}(0, I)$
        \State $\begin{aligned}[t]
           \pi \;\leftarrow\; &\pi
           + \sigma_k^{2}\,\eta\,\nabla \log p_{\sigma_k}(\pi) \\
           &{}+ \sqrt{\,2\,\sigma_k^{2}\,\eta\,\tau\,}\;\xi_k
        \end{aligned}$
    \State $\pi_i^0 \leftarrow s_i,\quad \pi_i^H \leftarrow g_i \quad \forall\, i$
    \EndFor
    \State \textbf{Return:} refined trajectory $\pi$
    \end{algorithmic}
\end{algorithm}
\end{minipage}
\vspace{-0.5\baselineskip}
\end{wrapfigure}
Where $d(\pi_i^h, \mathcal{C}_{\mathrm{f}})$ is a signed distance function (being the negative distance to $\mathcal{C} \setminus \mathcal{C}_{\mathrm{f}}$ if $\pi_i^h \in \mathcal{C}_{\mathrm{f}}$). 
We instantiate $\psi^{\sigma}$ on two feasible sets derived from
Eqs.~\eqref{eq:local-C-agent}--\eqref{eq:local-C-dyn}: the static obstacle-free
region $\mathcal{C}_f^{\mathrm{obs}}(\pi_i^h) := \mathcal{C}_{\mathrm{obs}}(\pi_i^h;\mathcal{O}) = \mathcal{C}_f$,
and the agent-separation feasible region
$\mathcal{C}_f^{\mathrm{agent}}(\pi_i^h;\, \pi_{j\neq i}^h) := \mathcal{C}_{\mathrm{agent}}(\pi_i^h;\, \pi_{j\neq i}^h)$.
The corresponding constraint score terms used in
Eq.~\eqref{eq:local-score} are
\begin{align*}
\nabla \log \obsTerm{\chi_{\mathrm{obs}}^{\sigma}(\pi_i^h)}
&= \tfrac{1}{\sigma^2}\, \psi^{\sigma}\!\left(\pi_i^h,\, \mathcal{C}_f^{\mathrm{obs}}\right),
\\
\nabla \log \agentTerm{\chi_{\mathrm{agent}}^{\sigma}(\pi_i^h)}
&= \tfrac{1}{\sigma^2}\, \psi^{\sigma}\!\left(\pi_i^h,\, \mathcal{C}_f^{\mathrm{agent}}\right).
\label{eq:score-terms}
\end{align*}
The complete procedure is summarized in Algorithm~\ref{alg:analytical-diffusion-planner}.
An illustration of the true score function, our approximation, and the gradient given by SDF guidance is shown in Figure~\ref{fig:sdf_vs_diffusion_score}. 
Notice that the score pushes agents out of obstacles in proportion to their penetration depth. It also pushes agents apart only when their uncertainty neighborhoods overlap or approach collision. This gives a continuous analogue of conflict resolution in MAPF, but without constructing a discrete high-level search tree. 
We encode velocity constraints through a soft penalty:
\begin{equation*}
    U_{\mathrm{dyn}}(\pi)
    =
    \lambda_{\mathrm{vel}}
    \sum_i
    \sum_{h=0}^{H-1}
    \phi\left(\|\pi_i^h - \pi_i^{h + 1}\|-v_{\max}\right),
\end{equation*}
where $\phi(r)=\max\{0,r\}^2$ or a smooth approximation of the squared hinge. This discourages pairwise step displacements that exceed the velocity limit. We then set $\dynTerm{s_{\mathrm{dyn}}(\pi_i^h)} = -\nabla_{\pi_i^h} U_{\mathrm{dyn}}(\pi)$.

\begin{figure}[t]
    \centering
    \begin{subfigure}[t]{0.24\textwidth}
        \centering
        \includegraphics[width=\linewidth]{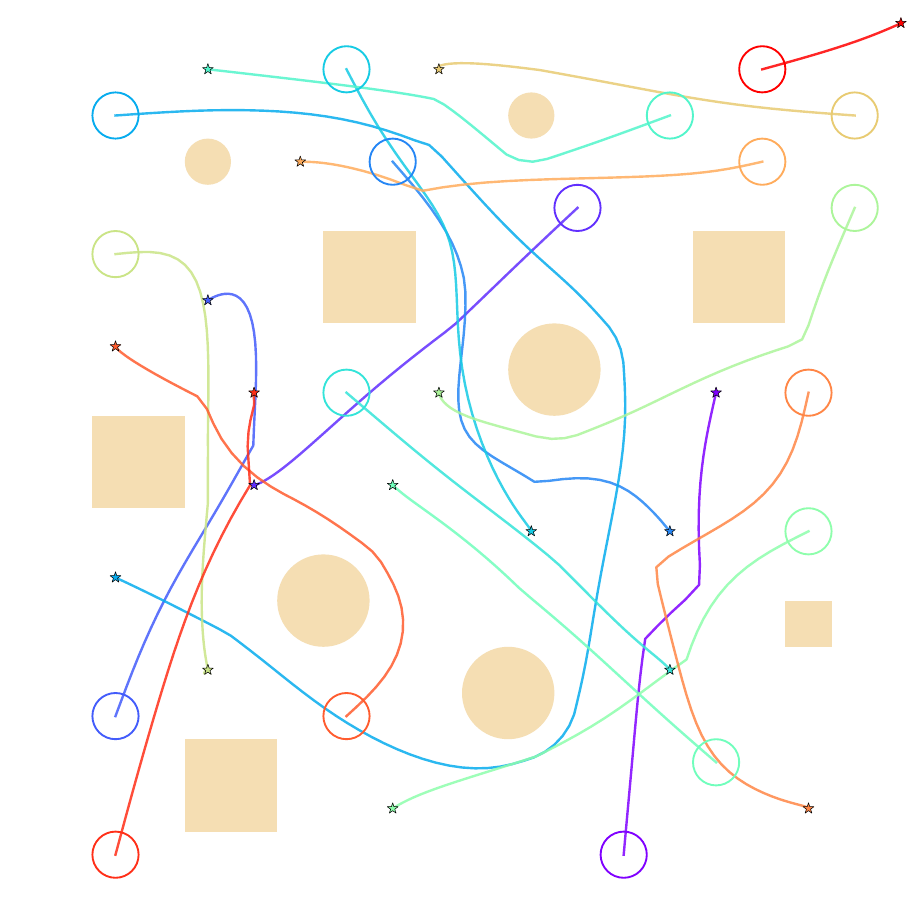}
        \caption{Basic}
    \end{subfigure}
    \hfill
    \begin{subfigure}[t]{0.24\textwidth}
        \centering
        \includegraphics[width=\linewidth]{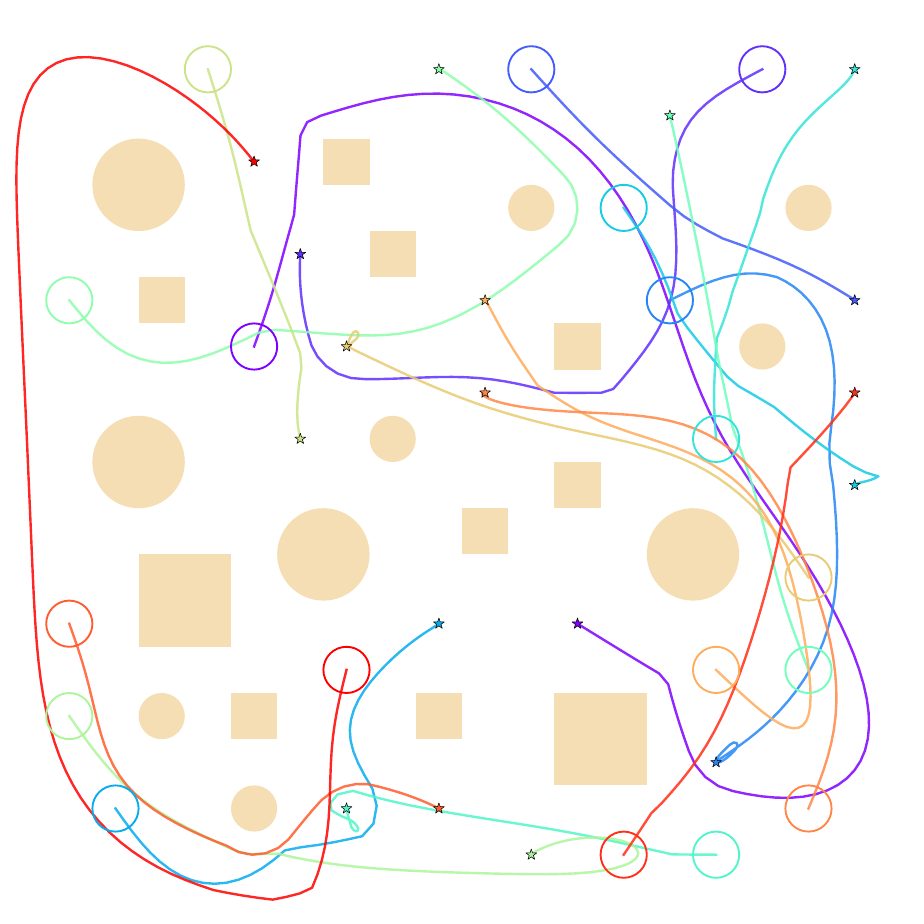}
        \caption{Dense}
    \end{subfigure}
    \hfill
    \begin{subfigure}[t]{0.24\textwidth}
        \centering
        \includegraphics[width=\linewidth]{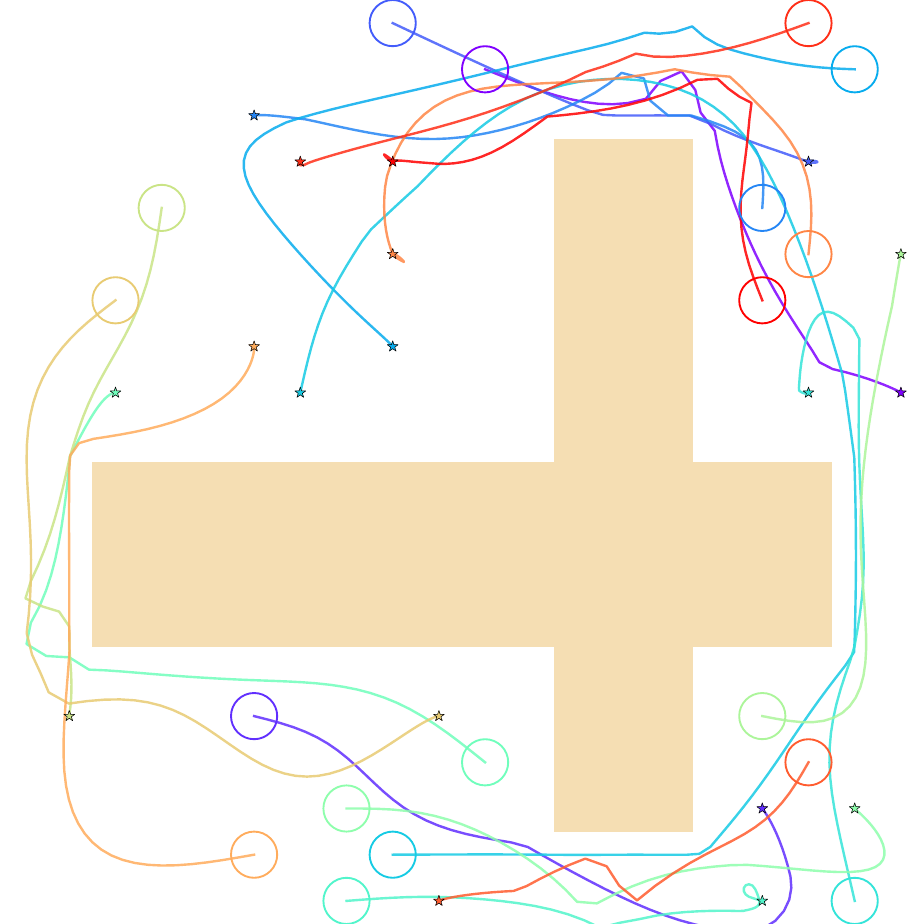}
        \caption{Room}
    \end{subfigure}
    \hfill
    \begin{subfigure}[t]{0.24\textwidth}
        \centering
        \includegraphics[width=\linewidth]{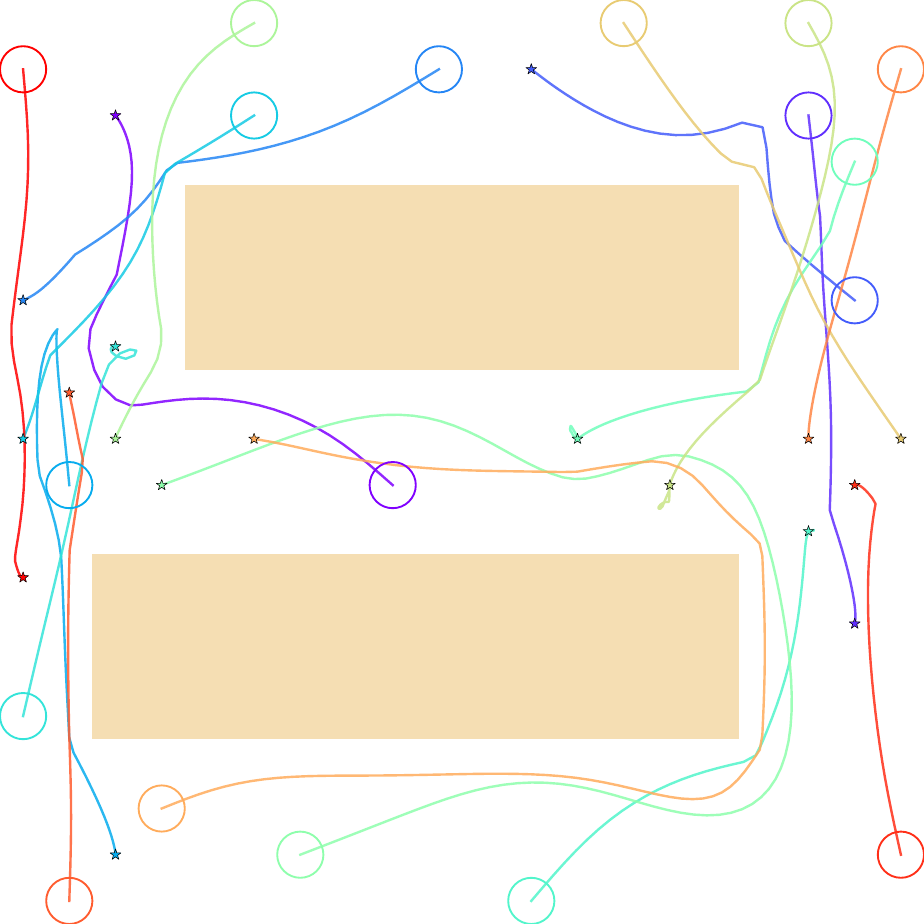}
        \caption{Shelf}
    \end{subfigure}
    \caption{Representative 18-robot trajectories generated by TFDP in four benchmark environments.}
    \label{fig:tfdp_18_robot_examples}
\end{figure}

\section{Experiments}

\textbf{Settings.}
To evaluate the performance of various multi-agent path planning algorithms, we design two experimental settings targeting different aspects of the problem:
\\
\textbf{1. Feasibility, Efficiency, Quality:} We adopt a recent multi-agent path planning benchmark~\citep{liang_discrete-guided_2025} which includes four representative map types: random maps with varying obstacle densities (basic and dense), and structured maps mimicking real-world environments (room and shelf layouts). For each map type, we consider three robot counts (6, 12, and 18), and evaluate across 25 instances with different configurations. Figure~\ref{fig:tfdp_18_robot_examples} illustrates the benchmark maps and representative $18$-robot trajectories found by our method.
\\
\textbf{2. Scalability:} To test scalability, we use a large-scale map containing over 100 heterogeneous obstacles. This setup is challenging and supports scenarios with up to 300 robots, largely beyond what existing diffusion methods handle. 

\textbf{Evaluation Metrics.}
Different algorithms are empirically evaluated based on \underline{Success rate}, \underline{Running time}, and \underline{Path Length}. The \emph{Success rate} indicates the proportion of test cases solved without collisions and within the time limit (900 seconds), \emph{Running time} measures the computational efficiency required to generate a solution, \emph{Path Length} evaluates the quality of generated trajectories.
\\[4pt]
\textbf{Competing Methods.}
Our proposed {\bf TFDP} is compared against diffusion-based and classical baseline methods:
    {(1) \textbf{Multi-robot motion planning diffusion (MMD)}: A method that integrates diffusion models with a classical search-based MAPF algorithm to generate multi-agent trajectories under soft collision constraints~\citep{shaoul_multi-robot_2025}.}
    {(2) \textbf{Simultaneous multi-robot motion planning diffusion (SMD)}: It integrates diffusion models with constrained optimization to generate collision-free trajectories for multi-agent path planning~\citep{liang_multi-agent_2024}.}
    {(3) \textbf{Discrete-guided diffusion (DGD)}: This approach decomposes the original multi-agent path planning problem and leverages MAPF-derived spatiotemporal dependencies to guide diffusion models in generating trajectories that satisfy hard feasibility constraints.~\citep{liang_discrete-guided_2025}.}
    {(4) \textbf{Model Predictive Path Integral (MPPI)}: A classical sampling-based method whose updates admit a score-ascent interpretation~\citep{xue2024fullordersamplingbasedmpctorquelevel}. This connection motivates its use as a baseline: MPPI estimates update directions from sampled perturbations, while TFDP uses analytical approximations to local scores.}
For MMD, SMD, and DGD, we follow the hyperparameter settings reported in their original papers. For MPPI, we treat the MPPI $\sigma$ the same as diffusion $\sigma$ and use the our approach's annealed energy landscapes as the optimization objective. For our method, we use $250$ Langevin steps, with an exponential noise schedule decreasing from $\sigma=0.1$ to $\sigma=0.0001$, and an exponentially annealed temperature from $\tau=0.3$ to $\tau=0.0001$. We also exponentially decrease the weight of the kinetic-energy term from $0.5$ to $0.001$ during sampling. The step size is $\eta = 0.5$. %

\vspace{-4pt}
\subsection{Comparison across Methods}
Table \ref{tab:results} presents the performance of TFDP and baseline methods across the four benchmark maps. The results show that TFDP consistently achieves a $100\%$ success rate in all but one configuration, matching or exceeding the success rates of the diffusion baselines.
In particular, on Dense with $18$ robots, TFDP achieves $96\%$ success, exceeding DGD's $76\%$ and MMD's $8\%$. Compared with DGD, the strongest diffusion baseline in these experiments, TFDP matches success in $8$ configurations and improves it in the remaining $4$. MMD performs well on Basic but its success rate drops sharply on Dense, Room, and Shelf. SMD reaches $100\%$ success with $6$ robots, but its computational cost limits evaluation to this setting. Appendix~\ref{appendix:confidence_intervals} reports $95\%$ Wilson intervals for TFDP success and $\pm 2$ standard errors for runtime and path length (Table~\ref{tab:tfdp-ci}).

In terms of running time, TFDP is significantly faster than all other diffusion-based methods. Across the $12$ configurations, TFDP requires $0.18$--$1.24$ seconds, averaging $0.59$ seconds, whereas DGD requires $12.2$--$324.8$ seconds, averaging $88.23$ seconds. The ratio of mean runtimes is approximately $150\times$, with TFDP also achieving higher average success. The difference is especially pronounced on Shelf with $18$ robots: TFDP requires $0.84$ seconds, versus $324.8$ seconds for DGD at $84\%$ success. MMD and SMD are also substantially slower and do not match TFDP's combination of feasibility and scalability. MPPI remains slower under the same annealing schedule, \emph{even when run on a GPU}. Running MPPI on the same CPU as TFDP is slower still, as reported in Appendix~\ref{appendix:mppi_cpu} (Table~\ref{tab:mppi_cpu_runtime}).

\begin{table}[t]
\centering
\footnotesize
\setlength{\tabcolsep}{3.25pt}
\setlength{\aboverulesep}{0pt}
\setlength{\belowrulesep}{0pt}
\begin{tabular}{ll|>{\columncolor{sidgray}}c>{\columncolor{sidgray}}c>{\columncolor{sidgray}}c|ccc|ccc|ccc|ccc}
\toprule
Map & Robots & \multicolumn{3}{>{\columncolor{sidgray}}c|}{TFDP} & \multicolumn{3}{c|}{DGD} & \multicolumn{3}{c|}{MMD} & \multicolumn{3}{c}{SMD} & \multicolumn{3}{c}{MPPI} \\
 & & S & T & P & S & T & P & S & T & P & S & T & P & S & T & P \\
\midrule
\multirow{3}{*}{\rotatebox[origin=c]{90}{\textbf{Basic}}}
 & 6  & \textbf{100} & \textbf{0.18} & \textbf{1.07} & \textbf{100} & 12.2 & 1.21 & \textbf{100} & 27.1 & 1.12 & \textbf{100} & 254.3 & 1.10 & \textbf{100} & 0.63 & 1.13 \\
 & 12 & \textbf{100} & \textbf{0.29} & \textbf{1.08} & 96 & 25.0 & 1.28 & \textbf{100} & 56.0 & 1.12 & - & - & - & \textbf{100} & 0.72 & 1.16 \\
 & 18 & \textbf{100} & \textbf{0.45} & \textbf{1.08} & 96 & 65.7 & 1.33 & 96 & 86.3 & 1.13 & - & - & - & 99 & 0.81 & 1.19 \\
\hline
\multirow{3}{*}{\rotatebox[origin=c]{90}{\textbf{Dense}}}
 & 6  & \textbf{100} & \textbf{0.35} & 1.26 & \textbf{100} & 12.3 & 1.26 & 40 & 36.5 & 1.15 & \textbf{100} & 287.3 & \textbf{1.13} & 99 & 0.85 & 1.19 \\
 & 12 & \textbf{100} & \textbf{0.66} & 1.29 & \textbf{100} & 32.9 & 1.35 & 8 & 62.5 & \textbf{1.15} & - & - & - & 99 & 1.09 & 1.46 \\
 & 18 & \textbf{96} & \textbf{1.06} & 1.32 & 76 & 47.8 & 1.37 & 8 & 87.1 & \textbf{1.18} & - & - & - & 94 & 1.40 & 1.59 \\
\hline
\multirow{3}{*}{\rotatebox[origin=c]{90}{\textbf{Room}}}
 & 6  & \textbf{100} & \textbf{0.43} & 1.41 & \textbf{100} & 38.9 & 1.44 & 24 & 55.5 & \textbf{1.16} & \textbf{100} & 181.0 & 1.19 & \textbf{100} & 0.93 & 1.57 \\
 & 12 & \textbf{100} & \textbf{0.78} & 1.45 & \textbf{100} & 84.0 & 1.50 & 4 & 122.6 & \textbf{1.14} & - & - & - & \textbf{100} & 1.31 & 1.69 \\
 & 18 & \textbf{100} & \textbf{1.24} & \textbf{1.45} & \textbf{100} & 219.0 & 1.55 & 0 & - & - & - & - & - & 99 & 1.66 & 1.77 \\
\hline 
\multirow{3}{*}{\rotatebox[origin=c]{90}{\textbf{Shelf}}}
 & 6  & \textbf{100} & \textbf{0.27} & 1.23 & \textbf{100} & 64.6 & 1.30 & 32 & 48.5 & 1.18 & \textbf{100} & 274.3 & \textbf{1.16} & \textbf{100} & 0.74 & 1.46 \\
 & 12 & \textbf{100} & \textbf{0.51} & 1.25 & \textbf{100} & 131.5 & 1.33 & 4 & 95.1 & \textbf{1.22} & N/A & N/A & N/A & \textbf{100} & 1.00 & 1.54 \\
 & 18 & \textbf{100} & \textbf{0.84} & \textbf{1.26} & 84 & 324.8 & 1.38 & 0 & N/A & N/A & N/A & N/A & N/A & 98 & 1.24 & 1.62 \\
\bottomrule
\end{tabular}
\vspace{-4pt}
\caption{Success rate (S$\uparrow$, in \%), running time (T$\downarrow$, in seconds), and path length (P$\downarrow$). Arrows indicate whether higher ($\uparrow$) or lower ($\downarrow$) values are better. `-' indicates that the method did not complete any instances within the time limit. MPPI times are for \textit{GPU}; rest are CPU.}
\label{tab:results}
\vspace{-12pt}
\end{table}

Finally, TFDP matches or improves DGD's path length in all $12$ configurations and produces strictly shorter paths in $11$. On Room and Shelf, these shorter paths accompany comparable feasibility and substantially lower runtime. MMD and SMD sometimes produce shorter paths, but path length must be interpreted jointly with success and evaluation coverage: MMD often fails on cluttered or structured maps, while SMD is evaluated only with $6$ robots. TFDP also tends to produce shorter paths than MPPI; the latter's noisier update is a possible explanation, although this mechanism is not isolated here. %
Representative $18$-robot trajectories appear in Figure~\ref{fig:tfdp_18_robot_examples}.

\Needspace{8\baselineskip}
\vspace{-4pt}
\subsection{Ablation Studies and Scalability}
\label{sec:sdf_guidance_ablation}
\vspace{-4pt}

To assess the contribution of the analytical scores, we replace them with CHOMP-style SDF penalty guidance while keeping TFDP's denoising schedule~\citep{zucker_chomp_2013,carvalho_motion_2023,shaoul_multi-robot_2025}. We vary the Langevin step size $\eta\in\{0.1,0.2,0.5\}$ and SDF scaling factor $\alpha\in\{0.05,0.1\}$. Note that, on Dense with $18$ robots, TFDP reaches $96\%$ success versus $84\%$ for the best displayed SDF setting (see Table~\ref{tab:sdf_comparison}), thus supporting an advantage especially in obstacle-dense settings. Appendix~\ref{appendix:hyperparameter_sweep} reports a broader sweep of both methods with Wilson intervals while Figure~\ref{fig:optimization_timeline} in Appendix~\ref{appendix:optimization_timelines} compares their refinement processes.
\par
\Needspace{20\baselineskip}

\begin{wrapfigure}{r}{0.48\linewidth}
\setlength{\intextsep}{6pt}
\begin{minipage}{\linewidth}
\centering
\scriptsize
\setlength{\tabcolsep}{2pt}
\setlength{\aboverulesep}{0pt}
\setlength{\belowrulesep}{0pt}
\resizebox{\linewidth}{!}{%
\begin{tabular}{ll|cc|cc|cc|>{\columncolor{sidgray}}c}
\toprule
Map & Robots
& \multicolumn{6}{c|}{SDF guidance $(\eta,\alpha)$}
& TFDP \\
 & 
& \multicolumn{2}{c|}{$\eta=.1$}
& \multicolumn{2}{c|}{$\eta=.2$}
& \multicolumn{2}{c|}{$\eta=.5$}
& $\eta=.5$ \\
 & 
& $.05$ & $.10$
& $.05$ & $.10$
& $.05$ & $.10$
& -- \\
\midrule
\multirow{3}{*}{\rotatebox[origin=c]{90}{\textbf{Basic}}}
 & 6  & \textbf{100} & \textbf{100} & \textbf{100} & \textbf{100} & \textbf{100} & \textbf{100} & \textbf{100} \\
 & 12 & \textbf{100} & 92  & \textbf{100} & \textbf{100} & 96  & 88  & \textbf{100} \\
 & 18 & 76  & 80  & 92  & 96  & 92  & 92  & \textbf{100} \\
\hline
\multirow{3}{*}{\rotatebox[origin=c]{90}{\textbf{Dense}}}
 & 6  & 92  & 92  & \textbf{100} & 96  & \textbf{100} & 96  & \textbf{100} \\
 & 12 & 84  & 84  & 96  & 92  & 84  & 88  & \textbf{99} \\
 & 18 & 72  & 52  & 76  & 76  & 84  & 80  & \textbf{96} \\
\hline
\multirow{3}{*}{\rotatebox[origin=c]{90}{\textbf{Room}}}
 & 6  & 92  & 96  & \textbf{100} & \textbf{100} & \textbf{100} & \textbf{100} & \textbf{100} \\
 & 12 & 64  & 76  & 80  & 96  & 96  & 96  & \textbf{99} \\
 & 18 & 44  & 20  & 76  & 64  & 84  & 84  & \textbf{98} \\
\hline
\multirow{3}{*}{\rotatebox[origin=c]{90}{\textbf{Shelf}}}
 & 6  & 92  & 96  & \textbf{100} & \textbf{100} & \textbf{100} & \textbf{100} & \textbf{100} \\
 & 12 & 76  & 76  & 92  & 88  & \textbf{100} & \textbf{100} & \textbf{100} \\
 & 18 & 56  & 52  & 92  & 76  & 88  & 92  & \textbf{100} \\
\bottomrule
\end{tabular}}
\captionof{table}{Success rates (\%) for SDF guidance and TFDP.}
\label{tab:sdf_comparison}
\end{minipage}
\end{wrapfigure}

Next, recall that the analytical scores are local in time, depending only on a short temporal neighborhood of each waypoint. This locality is important for scalability as it allows TFDP to refine each agent's trajectory independently, without constructing a joint high-dimensional search space. %
We assess locality by training a neural score estimator with different temporal receptive fields. We notice that, at low noise levels, prediction error plateaus as the window includes more distant timesteps, thus supporting the use of short temporal neighborhoods for refinement. For example, at $\sigma=0.05$, increasing the window from $1$ to $3$ reduces mean squared error from $0.8679$ to $0.3849$, while extending it to $7$ yields $0.2988$. Appendix~\ref{app:nn_score_dependence_test} reports the full results. Note that we use this estimator used only to test locality; TFDP remains training-free.
\\[4pt]
Next, we ablate the coarse planner used to initialize TFDP. We compare RRT initialization with Voronoi roadmaps, straight-line interpolation, and Brownian bridges between starts and goals. {On Room and Shelf, straight-line and Brownian-bridge initialization achieves only $0$--$17\%$ success, versus $99$--$100\%$ for Voronoi and RRT. Thus, initialization is critical in these structured layouts, but the ones adopted are also extremely easy and cheap to compute.}
Appendix~\ref{appendix:coarse_planner} reports all four initializations for both TFDP and MPPI.
\\[4pt]
Finally, we compare the default noise schedule with noiseless, deterministic annealed refinement. Removing noise matches or exceeds noisy success on the standard maps, whereas noise improves success point estimates by $5$--$7$ percentage points in the crowded $8$- and $12$-agent Bottleneck and Standoff cases (Appendix~\ref{appendix:effect_of_noise}). Thus, among the tested settings, noise is most useful in the more constrained scenarios. It can also improve a weak initialization: with straight-line trajectories replacing RRT, {suitable choices of temperature $T$ and smoothing scale $\sigma$ can improve success}. Appendix~\ref{appendix:sigma_and_t} reports this ablation on Basic and Dense, showing that the benefit depends on the sampling parameters.

{\bf Scalability.} 
Next, we test TFDP on a large-scale map with $100$ obstacles and up to $300$ robots. The tested map is a $100 \times 100$ unit square with randomly placed obstacles of varying sizes. We generate random start and goal positions for each robot, ensuring that they are not in collision with obstacles or each other.
Importantly, TFDP produces valid plans for such maps in $5.39$ seconds, on average on an A100 GPU. %
None of the diffusion baselines tested were able to handle this scale, and MPPI fails to find feasible plans for more than $50$ robots. Figure~\ref{fig:bigmap_plan} in Appendix~\ref{appendix:planning_visualizations} shows a generated plan.
\\[4pt]
We furthertest two \emph{deadlock scenarios} in which robots must pass through a wall opening only wide enough for one robot (see Appendix~\ref{appendix:deadlock_prone}, Figure~\ref{fig:stress_test_maps}). Here, TFDP achieves much higher success than MPPI across the tested robot counts, indicating an advantage when passage through a shared bottleneck requires coordination.

\section{Conclusion}

This paper presented a training-free analytical diffusion planner for path finding and multi-agent path planning. The method replaces learned full-trajectory scores with decomposed analytical scores derived from smoothness, obstacle, velocity, and inter-agent feasibility terms. The analysis and experiments show that local temporal neighborhoods capture most of the useful score information needed for denoising, enabling efficient trajectory refinement without demonstrations or map-specific training. Future work will extend the method to richer kinodynamic models, more challenging maze-like environments, and large-scale multi-agent benchmarks.
\\
Note that, although this is standard to test multi-agent planning on 2D environments, it is interesting to consider extensions to higher dimensions, which we believe will be a natural next step. We also assume a centralized planner and perfect plan executions, and so it is interesting to consider how approaches involving diffusion-like sampling processes could be extended to decentralized planning problems.

\clearpage
\subsection*{AI Use Statement}
Generative AI assisted with the creation of Figure~\ref{fig:overview}, its caption,
and the associated manuscript edits. The figure is a schematic illustration of
the method. The authors also used AI to proofread the manuscript and suggest edits for clarity and conciseness. The authors verified all AI-generated content for accuracy and appropriateness, and they take full responsibility for the final content of the paper.

\subsection*{Ethics statement}
The research presented in this paper is focused on improving multi-agent path planning algorithms. The methods and experiments are designed to advance the state of the art in robotics and autonomous systems, with potential applications in areas such as warehouse automation, autonomous vehicles, and robotic swarms. The research is conducted with a commitment to transparency, reproducibility, and responsible use of technology.

\subsection*{Reproducibility statement}

The paper describes the problem formulation, analytical score construction,
planning algorithm, evaluation metrics, and sampling hyperparameters.
The appendix provides supporting mathematical derivations, additional
benchmark descriptions, hyperparameter sweeps, initialization and noise
ablations, and confidence intervals for the reported results.
The complete source code is available upon request and will be released
publicly upon acceptance of the paper.

\section*{Acknowledgments}
This research was supported by NSF grants 2334936, 2334448, NSF CAREER Award 2401285, and DARPA Contract No. \#HR0011252E005. The authors acknowledge Research Computing at the University of Virginia for providing computational resources that have contributed to the results reported within this paper. The views and conclusions contained in this document are those of the authors and should not be interpreted as representing the official policies, either expressed or implied, of the sponsoring organizations, agencies, or the U.S. government.

\bibliographystyle{iclr2027_conference}
\bibliography{references_editable}

\clearpage
\appendix
\raggedbottom

\section{Technical appendices and supplementary material}

\subsection{Conditional Information Flow}
\label{appendix:conditional_information_flow_derivation}

We start with the definition of \textit{mutual information} between a pair of random variables:

Particularly, for a bivariate normal distribution $p(x_i, x_j) = \mathcal{N}(x_i, x_j; 0, \Sigma)$,
\[
I(x_i; x_j) = -\frac{1}{2} \log \left(1 - \frac{\Sigma_{ij}^2}{\Sigma_{ii} \Sigma_{jj}} \right).
\]
Noting that $p(x_i, x_j \mid x_{k \neq i,j})$ has a precision matrix equal to a \textit{submatrix} of the original for all $x_{k \neq i,j}$, we can read off the mutual information from the precision matrix of the overall distribution. The conditional mutual information, in terms of the precision matrix, is
\[
I(x_i; x_j \mid x_{-\{i, j\}}) = -\frac{1}{2} \log \left(1 - \frac{(\Sigma^{-1}_{ij})^2}{\Sigma^{-1}_{jj} \Sigma^{-1}_{ii}}\right).
\]
If there is no connection between the two variables in the MRF (i.e., $(\Sigma^{-1})_{ij} = 0$), the variables are independent conditioned on all other variables. Starting from a precision matrix $\Sigma^{-1}$, with nonzero entries representing adjacent random variables in an MRF, one can determine the covariance of the resulting distribution after a forward diffusion process as $\Sigma_t = \Sigma + t I$. In terms of the precision matrix, this is $\Omega_t = (\Sigma + tI)^{-1}$. For very small $t$, $\Omega_t \approx \Omega - \Omega (tI) \Omega$, and
\[
\left. \frac{d}{dt} \Omega_t \right|_{t=0} = -\Omega^2
\]
Thus, if we have a Gaussian distribution given by $\Omega$, the change after an infinitesimal amount of diffusion time is $-\Omega^2$. When $\Omega$ is treated as an adjacency matrix, $\Omega^2$ represents the adjacency after two ``hops''. Thus, the net effect is that codependence between random variables spreads out across the MRF.

\subsection{Analytical Score Function Derivation}
\label{appendix:analytical_score_derivation}

For general energy landscapes $\log p(x)$, we note that the ``score'' of the landscape convolved with $\mathcal{N}(0, \sigma^2 I)$ can be computed as the quotient of two integrals:
\begin{equation}
\label{eqn:score_integral}
\nabla_x \log p_{\sigma}(x) = \frac{1}{\sigma^2} \frac{\int (\tilde{x} - x) \mathcal{N}(\tilde{x}; 0, \sigma^2 I) p(\tilde{x}) \dif \tilde{x}}{\int \mathcal{N}(\tilde{x}; 0, \sigma^2 I) p(\tilde{x}) \dif \tilde{x}}
\end{equation}
We then provide methods for computing these integrals for multi-agent pathfinding constraint types.

\phantomsection
\label{appendix:circular_obstacle_score_derivation}
\textbf{Circular exclusion constraints}. These represent noncollision constraints between circular agents and circular obstacles or other agents. We assume the agent is positioned at $(0, 0) \in \R^2$, with radius $r := r_{agent}$, and that the obstacle is positioned at $(x_{obs}, 0) \in \R^2$ with effective radius $R := r_{obstacle} + r_{agent}$. The feasible region is $\mathcal{C} = \{(x, y) \in \R^2 \mid (x - x_{obs})^2 + y^2 > R^2\}$. Because the agent is at $(x, y) = (0, 0)$, we may replace $x \to 0$ in \ref{eqn:score_integral}.

Then let $r_1 = \min\, \{r > 0 \mid \exists x \in \delta \mathcal{C} : |x|^2 = r\}$ and $r_2 = \max\, \{r > 0 \mid \exists x \in \delta \mathcal{C} : |x|^2 = r\}$ (i.e., the minimum and maximum distances from the agent that are not wholly contained or wholly excluded from the object). Furthermore, let $\theta_{\text{min}}(r) = \min \, \{\theta \in [-\pi, \pi] \mid (r \cos \theta, r \sin \theta) \in \mathcal{C}\}$, $\theta_{\text{max}}(r) = \max \, \{\theta \in [-\pi, \pi] \mid (r \cos \theta, r \sin \theta) \in \mathcal{C}\}$. Since the coordinate system is such that the obstacle lies on the $x$ axis, $\theta_{min}(r) = -\theta_{max}(r)$. Then,
\begin{align}
\label{eqn:circular_exclusion_numerator_integral}
\int \tilde{x} \mathcal{N}(\tilde{x}; 0, \sigma^2 I) p(\tilde{x}) \dif \tilde{x}
&= \int_{r_1}^{r_2} r^2 \mathcal{N}(\tilde{x}; 0, \sigma^2 I) \int_{-\pi}^{\pi} (\cos \theta \, \vec{e}_1 + \sin \theta \, \vec{e}_2) \, \mathbb{I}_\mathcal{C}(\tilde{x}) \dif \theta \dif r \\
&\propto \int_{r_1}^{r_2} r^2 e^{-\frac{r^2}{2 \sigma^2}} \int_{\theta_{\text{min}}(r)}^{\theta_{\text{max}}(r)} \cos \theta \, \vec{e}_1 \dif \theta \dif r \\
&= \int_{r_1}^{r_2} 2 r^2 e^{-\frac{r^2}{2 \sigma^2}} \sin \theta_{\max}(r) \, \vec{e}_1 \dif r
\end{align}
and,
\begin{align}
\label{eqn:circular_exclusion_denominator_integral}
\int \mathcal{N}(\tilde{x}; 0, \sigma^2 I) p(\tilde{x}) \dif \tilde{x}
&\propto \int_{r_1}^{r_2} r e^{-\frac{r^2}{2 \sigma^2}} \int_{-\pi}^{\pi} \mathbb{I}_\mathcal{C}(\tilde{x}) \dif \theta \dif r \\
&= \int_{r_1}^{r_2} 2 r e^{-\frac{r^2}{2 \sigma^2}} \theta_{\max}(r) \dif r
\end{align}
To compute $\theta_{\max}$,
\begin{align*}
\begin{cases}
    x^2 + y^2 &= r^2 \\
    (x - x_{obs})^2 + y^2 &= R^2
\end{cases}
& \implies x^2 + y^2 - 2 x_{obs} x + x_{obs}^2 = R^2 \\
& \implies -2 x_{obs} x + x_{obs}^2 = R^2 - r^2 \\
& \implies x = -\frac{R^2 - r^2 - x_{obs}^2}{2 x_{obs}} \\
& \implies y = \sqrt{r^2 - x^2} = r \sin \theta_{\max}
\end{align*}
Then $r \sin \theta_{\max}$ may be used directly in \ref{eqn:circular_exclusion_numerator_integral}, and one can compute $\theta_{\max} = \text{arctan2}(y, x)$ for \ref{eqn:circular_exclusion_denominator_integral}.

\subsection{Rectangular Feasible Region Score Function Derivation}
\label{appendix:rectangular_box_score_function_derivation}

Consider a 2D region $\mathcal{C} = [x_{\min}, x_{\max}] \times [y_{\min}, y_{\max}]$. A uniform distribution over this region has the probability density:
\[
p(x, y) = \frac{\mathbb{I}_{\mathcal{C}}(x, y)}{(x_{\max} - x_{\min})(y_{\max} - y_{\min})}
\]
To create a diffusion-based generative model of this region, we can convolve this region with the Gaussian $\mathcal{N}(0, \sigma^2 I)$. Let $q = (x, y) \in \R^2$. Then the distribution at noise level $\sigma$ is given by:
\[
p_{\sigma}(q) = \int \mathcal{N}(q; \tilde{q}, \sigma^2 I) p(\tilde{q}) \dif \tilde{q}
\]
Because $p(\tilde{q})$ is proportional to the indicator function over the feasible set, $p_{\sigma}(q)$ is proportional to an integral over the feasible region:
\[
p_{\sigma}(q) \propto \int_{\mathcal{C}} \mathcal{N}(q; \tilde{q}, \sigma^2 I) \dif \tilde{q}
\]
This represents the overlap between a Gaussian distribution with mean $q = (x, y)$ and standard deviation $\sigma$ and the feasible region. Taking the score,
\[
\nabla \log p_\sigma(q) = \frac{\nabla p_\sigma(q)}{p_\sigma(q)}
\]
Since $\mathcal{C}$ is rectangular, we can express the denominator in terms of the Gaussian CDF, $\Phi(x) = \int_{-\infty}^{x} \mathcal{N}(\tilde{x}; 0, I) \dif \tilde{x}$.
\begin{align*}
& \int_{x_{\min}}^{x_{\max}} \int_{y_{\min}}^{y_{\max}} \mathcal{N}(q; \tilde{q}, \sigma^2 I) \dif \tilde{x} \\
&= \left(\int_{x_{\min}}^{x_{\max}} \mathcal{N}(x; \tilde{x}, \sigma^2 I) \dif \tilde{x}\right) \left( \int_{y_{\min}}^{y_{\max}} \mathcal{N}(y; \tilde{y}, \sigma^2 I) \dif \tilde{y} \right) \\
&= \left(\Phi\left(\frac{x_{\max} - x}{\sigma}\right) - \Phi\left(\frac{x_{\min} - x}{\sigma}\right)\right) \left(\Phi\left(\frac{y_{\max} - y}{\sigma}\right) - \Phi\left(\frac{y_{\min} - y}{\sigma}\right)\right)
\end{align*}
Similarly, the numerator can be computed through the following observation:
\[
\int_{x_{\min}}^{x_{\max}} \mathcal{N}(x; \tilde{x}, \sigma^2 I) \dif \tilde{x} = \int_{0}^{x_{\max}} \mathcal{N}(x; \tilde{x}, \sigma^2 I) \dif \tilde{x} - \int_{0}^{x_{\min}} \mathcal{N}(x; \tilde{x}, \sigma^2 I) \dif \tilde{x}
\]
By the fundamental theorem of calculus,
\[
\frac{\dif}{\dif x} \int_{x_{\min}}^{x_{\max}} \mathcal{N}(x; \tilde{x}, \sigma^2 I) \dif \tilde{x} = \mathcal{N}(x; x_{\max}, \sigma^2 I) - \mathcal{N}(x; x_{\min}, \sigma^2 I)
\]
Therefore,
\begin{align*}
\nabla p_{\sigma}(q) &= (\mathcal{N}(x; x_{\max}, \sigma^2 I) - \mathcal{N}(x; x_{\min}, \sigma^2 I)) \left(\Phi\left(\frac{y_{\max} - y}{\sigma}\right) - \Phi\left(\frac{y_{\min} - y}{\sigma}\right)\right) \vec{e}_x \\
&+ \left(\Phi\left(\frac{x_{\max} - x}{\sigma}\right) - \Phi\left(\frac{x_{\min} - x}{\sigma}\right)\right) (\mathcal{N}(y; y_{\max}, \sigma^2 I) - \mathcal{N}(y; y_{\min}, \sigma^2 I)) \vec{e}_y
\end{align*}
Finally,
\[
\nabla \log p_{\sigma}(q) = \frac{\mathcal{N}(x; x_{\max}, \sigma^2 I) - \mathcal{N}(x; x_{\min}, \sigma^2 I)}{\Phi\left(\frac{x_{\max} - x}{\sigma}\right) - \Phi\left(\frac{x_{\min} - x}{\sigma}\right)} \vec{e}_x + \frac{\mathcal{N}(y; y_{\max}, \sigma^2 I) - \mathcal{N}(y; y_{\min}, \sigma^2 I)}{\Phi\left(\frac{y_{\max} - y}{\sigma}\right) - \Phi\left(\frac{y_{\min} - y}{\sigma}\right)} \vec{e}_y
\]

\subsection{Rectangular Infeasible Region Score Function Derivation}
\label{appendix:rectangular_infeasible_region_score_function_derivation}

To compute the score function for a rectangular \textit{infeasible} region, we can reuse many of our prior results. Although technically, the resulting distribution is not normalizable, one can still plug in the equation and get a reasonable result. Let's say now that $\mathcal{C} = ([x_{\min}, x_{\max}] \times [y_{\min}, y_{\max}])^c =: \mathcal{C}_{\text{inclusion}}^c$, the complement of the feasible region used in the previous section (which we shall denote $\mathcal{C}_{\text{inclusion}}$). Then we adopt the perspective that $p_{\sigma}(x)$ is proportional to the overlap between a Gaussian distribution of mean $q = (x, y)$ and standard deviation $\sigma$ and the feasible region. We get:
\begin{align*}
\nabla p_{\sigma}(q)
&= \nabla \int_{\mathcal{C}} \mathcal{N}(\tilde{q}; q, \sigma^2 I) \dif \tilde{q} \\
&= \nabla \left( \int \mathcal{N}(\tilde{q}; q, \sigma^2 I) \dif \tilde{q} - \int_{\mathcal{C}_{\text{inclusion}}} \mathcal{N}(\tilde{q}; q, \sigma^2 I) \dif \tilde{q} \right) \\
&= \nabla \left( 1 - \int_{\mathcal{C}_{\text{inclusion}}} \mathcal{N}(\tilde{q}; q, \sigma^2 I) \dif \tilde{q} \right) \\
&= -\nabla \int_{\mathcal{C}_{\text{inclusion}}} \mathcal{N}(\tilde{q}; q, \sigma^2 I) \dif \tilde{q} \\
\end{align*}
Therefore, we simply negate the numerator calculation from \ref{appendix:rectangular_box_score_function_derivation}. For the denominator, we can use a similar trick by subtracting the denominator calculation from the previous section from $1$. The overall result is:
\begin{align*}
    \nabla \log p_{\sigma}(q) &= \frac{-(\mathcal{N}(x; x_{\max}, \sigma^2 I) - \mathcal{N}(x; x_{\min}, \sigma^2 I)) \left(\Phi\left(\frac{y_{\max} - y}{\sigma}\right) - \Phi\left(\frac{y_{\min} - y}{\sigma}\right)\right)}{1 - \left(\Phi\left(\frac{x_{\max} - x}{\sigma}\right) - \Phi\left(\frac{x_{\min} - x}{\sigma}\right)\right) \left(\Phi\left(\frac{y_{\max} - y}{\sigma}\right) - \Phi\left(\frac{y_{\min} - y}{\sigma}\right)\right)} \vec{e}_x\\
&+ \frac{-(\mathcal{N}(y; y_{\max}, \sigma^2 I) - \mathcal{N}(y; y_{\min}, \sigma^2 I)) \left(\Phi\left(\frac{x_{\max} - x}{\sigma}\right) - \Phi\left(\frac{x_{\min} - x}{\sigma}\right)\right)}{1 - \left(\Phi\left(\frac{x_{\max} - x}{\sigma}\right) - \Phi\left(\frac{x_{\min} - x}{\sigma}\right)\right) \left(\Phi\left(\frac{y_{\max} - y}{\sigma}\right) - \Phi\left(\frac{y_{\min} - y}{\sigma}\right)\right)} \vec{e}_y
\end{align*}

\subsection{Connection to Projection-Based Methods}
\label{appendix:projection_at_low_sigma}
The Stein score for a distribution $p_\sigma(x)$ is defined as $\nabla_x \log p_\sigma(x)$. This may be extended to a more elucidating form:
\begin{align*}
\nabla_x \log p_\sigma(x)
&= \frac{\nabla_x p_{\sigma}(x)}{p_{\sigma}(x)} \\
&= \frac{\int \nabla_x \mathcal{N}(x - \tilde{x}; 0, \sigma^2 I) p(\tilde{x}) \dif \tilde{x}}{p_\sigma(x)} \\
&= \frac{\int \mathcal{N}(x - \tilde{x}; 0, \sigma^2 I) \nabla_x \log \mathcal{N}(x - \tilde{x}; 0, \sigma^2 I) p(\tilde{x}) \dif \tilde{x}}{p_\sigma(x)} \\
&= \frac{1}{\sigma^2} \frac{\int_{\R^n} (\tilde{x} - x) \mathcal{N}(x - \tilde{x}; 0, \sigma^2 I) p(\tilde{x}) \dif \tilde{x}}{p_\sigma(x)}
\end{align*}
For linear inequality or box constraints, the necessary integrals may be computed analytically, and for other low-dimensional constraints, the integral may be computed numerically. Furthermore, we can note that happens for low-$\sigma$ behavior. If we assume $p(x) = p(x) = \frac{\mathbb{I}_\mathcal{C}(x)}{m(\mathcal{C})}$, then:
\begin{align*}
    \lim_{\sigma \to 0} \sigma^2 \nabla_x \log p_\sigma(x) &= \lim_{\sigma \to 0} \frac{\int (\tilde{x} - x) \mathcal{N}(x - \tilde{x}; 0, \sigma^2 I) p(\tilde{x}) \dif \tilde{x}}{\int \mathcal{N}(x - \tilde{x}; 0, \sigma^2 I) p(\tilde{x}) \dif \tilde{x}} \\
    &= \lim_{\sigma \to 0} \frac{\int (\tilde{x} - x) \exp\left(-\frac{(x - \tilde{x})^2}{2\sigma^2}\right) p(\tilde{x}) \dif \tilde{x}}{\int \exp\left(-\frac{(x - \tilde{x})^2}{2\sigma^2}\right) p(\tilde{x}) \dif \tilde{x}} \\
    &= \lim_{\sigma \to 0} \frac{\int_{\mathcal{C}} (\tilde{x} - x) \exp\left(-\frac{(x - \tilde{x})^2}{2\sigma^2}\right) \dif \tilde{x}}{\int_{\mathcal{C}} \exp\left(-\frac{(x - \tilde{x})^2}{2\sigma^2}\right) \dif \tilde{x}} \\
    &= \Pi_\mathcal{C}(x) - x
\end{align*}
That is, at very low $\sigma$, taking a discrete-time Langevin step with size $\sigma^2$ is precisely the projection operator:
\begin{align*}
x_{t+1}
&= x_t + \sigma^2 \nabla_x \log p_\sigma(x_t) + 2 \sigma z \\
&= x_t + (\Pi_\mathcal{C}(x_t) - x_t) + 2 \sigma z \\
&= \Pi_\mathcal{C}(x_t) + 2 \sigma z
\end{align*}
Since $\sigma \ll 1$, the Brownian noise term is negligible.

\subsection{Larger Hyperparameter Sweep}
\label{appendix:hyperparameter_sweep}

We vary the Langevin step size $\eta$ for TFDP and CHOMP-style SDF guidance, with SDF scaling factors $\alpha\in\{0.05,0.1,0.2,0.5\}$. The reported Wilson confidence intervals use $10$ trials of $30$ maps each. Gray columns identify TFDP, and arrows indicate whether higher ($\uparrow$) or lower ($\downarrow$) values are better. Bold marks the best comparable central values, including ties; the captions specify the comparison groups.

\Needspace{23\baselineskip}
{\bf TFDP, smaller steps.} Table~\ref{tab:sweep-tfdp-small} tests whether small updates suffice for reliable refinement. Success remains at $0.01$ for $\eta\leq0.02$, while $\eta=0.10$ raises it to $0.94$--$0.97$ with $6$ robots. The same step size reaches only $0.31$--$0.68$ with $18$ robots. Thus, the smaller-step regime is particularly restrictive as the number of interacting robots increases.

\begin{table}[H]
\centering
\footnotesize
\setlength{\tabcolsep}{3.25pt}
\setlength{\aboverulesep}{0pt}
\setlength{\belowrulesep}{0pt}
\begin{tabular}{ll|>{\columncolor{sidgray}}c>{\columncolor{sidgray}}c>{\columncolor{sidgray}}c>{\columncolor{sidgray}}c}
\toprule
Map & Robots & \multicolumn{4}{>{\columncolor{sidgray}}c}{Step size $\eta$} \\
 & & 0.01 & 0.02 & 0.05 & 0.10 \\
\midrule
\multirow{3}{*}{\rotatebox[origin=c]{90}{\textbf{Basic}}} & 6 & 0.01 $\pm$ 0.01 & 0.01 $\pm$ 0.01 & 0.31 $\pm$ 0.06 & \textbf{0.97} $\pm$ 0.02 \\
 & 12 & 0.01 $\pm$ 0.01 & 0.01 $\pm$ 0.01 & 0.07 $\pm$ 0.03 & \textbf{0.87} $\pm$ 0.04 \\
 & 18 & 0.01 $\pm$ 0.01 & 0.01 $\pm$ 0.01 & 0.01 $\pm$ 0.01 & \textbf{0.68} $\pm$ 0.06 \\
\hline
\multirow{3}{*}{\rotatebox[origin=c]{90}{\textbf{Dense}}} & 6 & 0.01 $\pm$ 0.01 & 0.01 $\pm$ 0.01 & 0.30 $\pm$ 0.06 & \textbf{0.96} $\pm$ 0.02 \\
 & 12 & 0.01 $\pm$ 0.01 & 0.01 $\pm$ 0.01 & 0.07 $\pm$ 0.03 & \textbf{0.84} $\pm$ 0.04 \\
 & 18 & 0.01 $\pm$ 0.01 & 0.01 $\pm$ 0.01 & 0.01 $\pm$ 0.01 & \textbf{0.56} $\pm$ 0.06 \\
\hline
\multirow{3}{*}{\rotatebox[origin=c]{90}{\textbf{Room}}} & 6 & 0.01 $\pm$ 0.01 & 0.01 $\pm$ 0.01 & 0.24 $\pm$ 0.05 & \textbf{0.94} $\pm$ 0.03 \\
 & 12 & 0.01 $\pm$ 0.01 & 0.01 $\pm$ 0.01 & 0.03 $\pm$ 0.02 & \textbf{0.69} $\pm$ 0.06 \\
 & 18 & 0.01 $\pm$ 0.01 & 0.01 $\pm$ 0.01 & 0.01 $\pm$ 0.01 & \textbf{0.31} $\pm$ 0.06 \\
\hline
\multirow{3}{*}{\rotatebox[origin=c]{90}{\textbf{Shelf}}} & 6 & 0.01 $\pm$ 0.01 & 0.01 $\pm$ 0.01 & 0.30 $\pm$ 0.06 & \textbf{0.96} $\pm$ 0.02 \\
 & 12 & 0.01 $\pm$ 0.01 & 0.01 $\pm$ 0.01 & 0.04 $\pm$ 0.02 & \textbf{0.75} $\pm$ 0.05 \\
 & 18 & 0.01 $\pm$ 0.01 & 0.01 $\pm$ 0.01 & 0.02 $\pm$ 0.02 & \textbf{0.61} $\pm$ 0.06 \\
\bottomrule
\end{tabular}
\caption{TFDP: success probability ($\uparrow$), with Wilson intervals, for smaller step sizes. Bold marks row maxima within this displayed range.}
\label{tab:sweep-tfdp-small}
\end{table}

\Needspace{23\baselineskip}
{\bf TFDP, larger steps.} Table~\ref{tab:sweep-tfdp-large} shows that larger updates substantially improve feasibility. At $\eta=0.50$, success is at least $0.95$ in every setting. Increasing $\eta$ further raises the Dense $18$-robot point estimate from $0.95$ to $0.97$, but lowers Room from $0.99$ to $0.94$ at $\eta=1.00$. This supports $\eta=0.50$ as a broadly effective choice, without identifying a universal optimum.

\begin{table}[H]
\centering
\footnotesize
\setlength{\tabcolsep}{3.25pt}
\setlength{\aboverulesep}{0pt}
\setlength{\belowrulesep}{0pt}
\begin{tabular}{ll|>{\columncolor{sidgray}}c>{\columncolor{sidgray}}c>{\columncolor{sidgray}}c>{\columncolor{sidgray}}c}
\toprule
Map & Robots & \multicolumn{4}{>{\columncolor{sidgray}}c}{Step size $\eta$} \\
 & & 0.20 & 0.50 & 0.80 & 1.00 \\
\midrule
\multirow{3}{*}{\rotatebox[origin=c]{90}{\textbf{Basic}}} & 6 & \textbf{0.99} $\pm$ 0.01 & \textbf{0.99} $\pm$ 0.01 & \textbf{0.99} $\pm$ 0.01 & \textbf{0.99} $\pm$ 0.01 \\
 & 12 & \textbf{0.99} $\pm$ 0.01 & \textbf{0.99} $\pm$ 0.01 & \textbf{0.99} $\pm$ 0.01 & \textbf{0.99} $\pm$ 0.01 \\
 & 18 & 0.97 $\pm$ 0.02 & \textbf{0.99} $\pm$ 0.01 & \textbf{0.99} $\pm$ 0.01 & \textbf{0.99} $\pm$ 0.01 \\
\hline
\multirow{3}{*}{\rotatebox[origin=c]{90}{\textbf{Dense}}} & 6 & \textbf{0.99} $\pm$ 0.01 & \textbf{0.99} $\pm$ 0.01 & \textbf{0.99} $\pm$ 0.01 & \textbf{0.99} $\pm$ 0.01 \\
 & 12 & 0.96 $\pm$ 0.02 & 0.98 $\pm$ 0.02 & \textbf{0.99} $\pm$ 0.01 & 0.98 $\pm$ 0.02 \\
 & 18 & 0.89 $\pm$ 0.04 & 0.95 $\pm$ 0.03 & \textbf{0.97} $\pm$ 0.02 & \textbf{0.97} $\pm$ 0.02 \\
\hline
\multirow{3}{*}{\rotatebox[origin=c]{90}{\textbf{Room}}} & 6 & \textbf{0.99} $\pm$ 0.01 & \textbf{0.99} $\pm$ 0.01 & \textbf{0.99} $\pm$ 0.01 & \textbf{0.99} $\pm$ 0.01 \\
 & 12 & 0.97 $\pm$ 0.02 & \textbf{0.99} $\pm$ 0.01 & 0.98 $\pm$ 0.02 & 0.98 $\pm$ 0.02 \\
 & 18 & 0.92 $\pm$ 0.03 & \textbf{0.99} $\pm$ 0.01 & 0.95 $\pm$ 0.03 & 0.94 $\pm$ 0.03 \\
\hline
\multirow{3}{*}{\rotatebox[origin=c]{90}{\textbf{Shelf}}} & 6 & \textbf{0.99} $\pm$ 0.01 & \textbf{0.99} $\pm$ 0.01 & \textbf{0.99} $\pm$ 0.01 & \textbf{0.99} $\pm$ 0.01 \\
 & 12 & 0.98 $\pm$ 0.01 & \textbf{0.99} $\pm$ 0.01 & \textbf{0.99} $\pm$ 0.01 & \textbf{0.99} $\pm$ 0.01 \\
 & 18 & 0.95 $\pm$ 0.03 & 0.98 $\pm$ 0.01 & \textbf{0.99} $\pm$ 0.01 & 0.98 $\pm$ 0.01 \\
\bottomrule
\end{tabular}
\caption{TFDP: success probability ($\uparrow$), with Wilson intervals, for larger step sizes. Bold marks row maxima within this displayed range.}
\label{tab:sweep-tfdp-large}
\end{table}

\Needspace{23\baselineskip}
{\bf SDF with $\alpha=0.1$, smaller steps.} Table~\ref{tab:sweep-sdf010-small} shows the interaction between step size and robot count. At $\eta=0.10$, Room success decreases from $0.94$ with $6$ robots to $0.31$ with $18$, while Basic decreases from $0.97$ to $0.68$. Even the best step size in this panel leaves a substantial feasibility gap as coordination becomes more demanding.

\begin{table}[H]
\centering
\footnotesize
\setlength{\tabcolsep}{3.25pt}
\setlength{\aboverulesep}{0pt}
\setlength{\belowrulesep}{0pt}
\begin{tabular}{ll|cccc}
\toprule
Map & Robots & \multicolumn{4}{c}{Step size $\eta$} \\
 & & 0.01 & 0.02 & 0.05 & 0.10 \\
\midrule
\multirow{3}{*}{\rotatebox[origin=c]{90}{\textbf{Basic}}} & 6 & 0.01 $\pm$ 0.01 & 0.01 $\pm$ 0.01 & 0.31 $\pm$ 0.06 & \textbf{0.97} $\pm$ 0.02 \\
 & 12 & 0.01 $\pm$ 0.01 & 0.01 $\pm$ 0.01 & 0.07 $\pm$ 0.03 & \textbf{0.87} $\pm$ 0.04 \\
 & 18 & 0.01 $\pm$ 0.01 & 0.01 $\pm$ 0.01 & 0.01 $\pm$ 0.01 & \textbf{0.68} $\pm$ 0.06 \\
\hline
\multirow{3}{*}{\rotatebox[origin=c]{90}{\textbf{Dense}}} & 6 & 0.01 $\pm$ 0.01 & 0.01 $\pm$ 0.01 & 0.30 $\pm$ 0.06 & \textbf{0.96} $\pm$ 0.02 \\
 & 12 & 0.01 $\pm$ 0.01 & 0.01 $\pm$ 0.01 & 0.07 $\pm$ 0.03 & \textbf{0.84} $\pm$ 0.04 \\
 & 18 & 0.01 $\pm$ 0.01 & 0.01 $\pm$ 0.01 & 0.01 $\pm$ 0.01 & \textbf{0.56} $\pm$ 0.06 \\
\hline
\multirow{3}{*}{\rotatebox[origin=c]{90}{\textbf{Room}}} & 6 & 0.01 $\pm$ 0.01 & 0.01 $\pm$ 0.01 & 0.24 $\pm$ 0.05 & \textbf{0.94} $\pm$ 0.03 \\
 & 12 & 0.01 $\pm$ 0.01 & 0.01 $\pm$ 0.01 & 0.03 $\pm$ 0.02 & \textbf{0.69} $\pm$ 0.06 \\
 & 18 & 0.01 $\pm$ 0.01 & 0.01 $\pm$ 0.01 & 0.01 $\pm$ 0.01 & \textbf{0.31} $\pm$ 0.06 \\
\hline
\multirow{3}{*}{\rotatebox[origin=c]{90}{\textbf{Shelf}}} & 6 & 0.01 $\pm$ 0.01 & 0.01 $\pm$ 0.01 & 0.30 $\pm$ 0.06 & \textbf{0.96} $\pm$ 0.02 \\
 & 12 & 0.01 $\pm$ 0.01 & 0.01 $\pm$ 0.01 & 0.04 $\pm$ 0.02 & \textbf{0.75} $\pm$ 0.05 \\
 & 18 & 0.01 $\pm$ 0.01 & 0.01 $\pm$ 0.01 & 0.02 $\pm$ 0.02 & \textbf{0.61} $\pm$ 0.06 \\
\bottomrule
\end{tabular}
\caption{SDF guidance with $\alpha=0.1$: success probability ($\uparrow$), with Wilson intervals, for smaller step sizes. Bold marks row maxima within this displayed range.}
\label{tab:sweep-sdf010-small}
\end{table}

\Needspace{23\baselineskip}
{\bf SDF with $\alpha=0.1$, larger steps.} Table~\ref{tab:sweep-sdf010-large} shows high success over several neighboring step sizes. Basic remains at $0.99$ for all robot counts when $\eta\geq0.50$, and Shelf remains between $0.98$ and $0.99$. Dense and Room retain greater sensitivity with $18$ robots, indicating that a broad effective range on simpler cases does not imply uniform robustness across layouts.

\begin{table}[H]
\centering
\footnotesize
\setlength{\tabcolsep}{3.25pt}
\setlength{\aboverulesep}{0pt}
\setlength{\belowrulesep}{0pt}
\begin{tabular}{ll|cccc}
\toprule
Map & Robots & \multicolumn{4}{c}{Step size $\eta$} \\
 & & 0.20 & 0.50 & 0.80 & 1.00 \\
\midrule
\multirow{3}{*}{\rotatebox[origin=c]{90}{\textbf{Basic}}} & 6 & \textbf{0.99} $\pm$ 0.01 & \textbf{0.99} $\pm$ 0.01 & \textbf{0.99} $\pm$ 0.01 & \textbf{0.99} $\pm$ 0.01 \\
 & 12 & \textbf{0.99} $\pm$ 0.01 & \textbf{0.99} $\pm$ 0.01 & \textbf{0.99} $\pm$ 0.01 & \textbf{0.99} $\pm$ 0.01 \\
 & 18 & 0.97 $\pm$ 0.02 & \textbf{0.99} $\pm$ 0.01 & \textbf{0.99} $\pm$ 0.01 & \textbf{0.99} $\pm$ 0.01 \\
\hline
\multirow{3}{*}{\rotatebox[origin=c]{90}{\textbf{Dense}}} & 6 & \textbf{0.99} $\pm$ 0.01 & \textbf{0.99} $\pm$ 0.01 & \textbf{0.99} $\pm$ 0.01 & \textbf{0.99} $\pm$ 0.01 \\
 & 12 & 0.96 $\pm$ 0.02 & 0.98 $\pm$ 0.02 & \textbf{0.99} $\pm$ 0.01 & 0.98 $\pm$ 0.02 \\
 & 18 & 0.89 $\pm$ 0.04 & 0.95 $\pm$ 0.03 & \textbf{0.97} $\pm$ 0.02 & \textbf{0.97} $\pm$ 0.02 \\
\hline
\multirow{3}{*}{\rotatebox[origin=c]{90}{\textbf{Room}}} & 6 & \textbf{0.99} $\pm$ 0.01 & \textbf{0.99} $\pm$ 0.01 & \textbf{0.99} $\pm$ 0.01 & \textbf{0.99} $\pm$ 0.01 \\
 & 12 & 0.97 $\pm$ 0.02 & \textbf{0.99} $\pm$ 0.01 & 0.98 $\pm$ 0.02 & 0.98 $\pm$ 0.02 \\
 & 18 & 0.92 $\pm$ 0.03 & \textbf{0.99} $\pm$ 0.01 & 0.95 $\pm$ 0.03 & 0.94 $\pm$ 0.03 \\
\hline
\multirow{3}{*}{\rotatebox[origin=c]{90}{\textbf{Shelf}}} & 6 & \textbf{0.99} $\pm$ 0.01 & \textbf{0.99} $\pm$ 0.01 & \textbf{0.99} $\pm$ 0.01 & \textbf{0.99} $\pm$ 0.01 \\
 & 12 & 0.98 $\pm$ 0.01 & \textbf{0.99} $\pm$ 0.01 & \textbf{0.99} $\pm$ 0.01 & \textbf{0.99} $\pm$ 0.01 \\
 & 18 & 0.95 $\pm$ 0.03 & 0.98 $\pm$ 0.01 & \textbf{0.99} $\pm$ 0.01 & 0.98 $\pm$ 0.01 \\
\bottomrule
\end{tabular}
\caption{SDF guidance with $\alpha=0.1$: success probability ($\uparrow$), with Wilson intervals, for larger step sizes. Bold marks row maxima within this displayed range.}
\label{tab:sweep-sdf010-large}
\end{table}

\Needspace{23\baselineskip}
{\bf SDF with $\alpha=0.5$, smaller steps.} Table~\ref{tab:sweep-sdf050-small} tests the largest SDF scale in the small-step regime. The highest success in every row occurs at $\eta=0.10$, while $\eta\leq0.02$ yields only $0.01$. The remaining variation at $\eta=0.10$, from $0.31$ on Room with $18$ robots to $0.97$ on Basic with $6$, shows that the same update scale can yield very different planning outcomes.

\begin{table}[H]
\centering
\footnotesize
\setlength{\tabcolsep}{3.25pt}
\setlength{\aboverulesep}{0pt}
\setlength{\belowrulesep}{0pt}
\begin{tabular}{ll|cccc}
\toprule
Map & Robots & \multicolumn{4}{c}{Step size $\eta$} \\
 & & 0.01 & 0.02 & 0.05 & 0.10 \\
\midrule
\multirow{3}{*}{\rotatebox[origin=c]{90}{\textbf{Basic}}} & 6 & 0.01 $\pm$ 0.01 & 0.01 $\pm$ 0.01 & 0.31 $\pm$ 0.06 & \textbf{0.97} $\pm$ 0.02 \\
 & 12 & 0.01 $\pm$ 0.01 & 0.01 $\pm$ 0.01 & 0.07 $\pm$ 0.03 & \textbf{0.87} $\pm$ 0.04 \\
 & 18 & 0.01 $\pm$ 0.01 & 0.01 $\pm$ 0.01 & 0.01 $\pm$ 0.01 & \textbf{0.68} $\pm$ 0.06 \\
\hline
\multirow{3}{*}{\rotatebox[origin=c]{90}{\textbf{Dense}}} & 6 & 0.01 $\pm$ 0.01 & 0.01 $\pm$ 0.01 & 0.30 $\pm$ 0.06 & \textbf{0.96} $\pm$ 0.02 \\
 & 12 & 0.01 $\pm$ 0.01 & 0.01 $\pm$ 0.01 & 0.07 $\pm$ 0.03 & \textbf{0.84} $\pm$ 0.04 \\
 & 18 & 0.01 $\pm$ 0.01 & 0.01 $\pm$ 0.01 & 0.01 $\pm$ 0.01 & \textbf{0.56} $\pm$ 0.06 \\
\hline
\multirow{3}{*}{\rotatebox[origin=c]{90}{\textbf{Room}}} & 6 & 0.01 $\pm$ 0.01 & 0.01 $\pm$ 0.01 & 0.24 $\pm$ 0.05 & \textbf{0.94} $\pm$ 0.03 \\
 & 12 & 0.01 $\pm$ 0.01 & 0.01 $\pm$ 0.01 & 0.03 $\pm$ 0.02 & \textbf{0.69} $\pm$ 0.06 \\
 & 18 & 0.01 $\pm$ 0.01 & 0.01 $\pm$ 0.01 & 0.01 $\pm$ 0.01 & \textbf{0.31} $\pm$ 0.06 \\
\hline
\multirow{3}{*}{\rotatebox[origin=c]{90}{\textbf{Shelf}}} & 6 & 0.01 $\pm$ 0.01 & 0.01 $\pm$ 0.01 & 0.30 $\pm$ 0.06 & \textbf{0.96} $\pm$ 0.02 \\
 & 12 & 0.01 $\pm$ 0.01 & 0.01 $\pm$ 0.01 & 0.04 $\pm$ 0.02 & \textbf{0.75} $\pm$ 0.05 \\
 & 18 & 0.01 $\pm$ 0.01 & 0.01 $\pm$ 0.01 & 0.02 $\pm$ 0.02 & \textbf{0.61} $\pm$ 0.06 \\
\bottomrule
\end{tabular}
\caption{SDF guidance with $\alpha=0.5$: success probability ($\uparrow$), with Wilson intervals, for smaller step sizes. Bold marks row maxima within this displayed range.}
\label{tab:sweep-sdf050-small}
\end{table}

\Needspace{23\baselineskip}
{\bf SDF with $\alpha=0.5$, larger steps.} Table~\ref{tab:sweep-sdf050-large} reports at least $0.95$ success in every setting at $\eta=0.50$. Larger steps offer little improvement on Basic and Shelf, while Room with $18$ robots declines to $0.94$ at $\eta=1.00$. The table supports a moderate step size across these cases; small differences among the larger-step estimates should be read together with their uncertainty intervals.

\begin{table}[H]
\centering
\footnotesize
\setlength{\tabcolsep}{3.25pt}
\setlength{\aboverulesep}{0pt}
\setlength{\belowrulesep}{0pt}
\begin{tabular}{ll|cccc}
\toprule
Map & Robots & \multicolumn{4}{c}{Step size $\eta$} \\
 & & 0.20 & 0.50 & 0.80 & 1.00 \\
\midrule
\multirow{3}{*}{\rotatebox[origin=c]{90}{\textbf{Basic}}} & 6 & \textbf{0.99} $\pm$ 0.01 & \textbf{0.99} $\pm$ 0.01 & \textbf{0.99} $\pm$ 0.01 & \textbf{0.99} $\pm$ 0.01 \\
 & 12 & \textbf{0.99} $\pm$ 0.01 & \textbf{0.99} $\pm$ 0.01 & \textbf{0.99} $\pm$ 0.01 & \textbf{0.99} $\pm$ 0.01 \\
 & 18 & 0.97 $\pm$ 0.02 & \textbf{0.99} $\pm$ 0.01 & \textbf{0.99} $\pm$ 0.01 & \textbf{0.99} $\pm$ 0.01 \\
\hline
\multirow{3}{*}{\rotatebox[origin=c]{90}{\textbf{Dense}}} & 6 & \textbf{0.99} $\pm$ 0.01 & \textbf{0.99} $\pm$ 0.01 & \textbf{0.99} $\pm$ 0.01 & \textbf{0.99} $\pm$ 0.01 \\
 & 12 & 0.96 $\pm$ 0.02 & 0.98 $\pm$ 0.02 & \textbf{0.99} $\pm$ 0.01 & 0.98 $\pm$ 0.02 \\
 & 18 & 0.89 $\pm$ 0.04 & 0.95 $\pm$ 0.03 & \textbf{0.97} $\pm$ 0.02 & \textbf{0.97} $\pm$ 0.02 \\
\hline
\multirow{3}{*}{\rotatebox[origin=c]{90}{\textbf{Room}}} & 6 & \textbf{0.99} $\pm$ 0.01 & \textbf{0.99} $\pm$ 0.01 & \textbf{0.99} $\pm$ 0.01 & \textbf{0.99} $\pm$ 0.01 \\
 & 12 & 0.97 $\pm$ 0.02 & \textbf{0.99} $\pm$ 0.01 & 0.98 $\pm$ 0.02 & 0.98 $\pm$ 0.02 \\
 & 18 & 0.92 $\pm$ 0.03 & \textbf{0.99} $\pm$ 0.01 & 0.95 $\pm$ 0.03 & 0.94 $\pm$ 0.03 \\
\hline
\multirow{3}{*}{\rotatebox[origin=c]{90}{\textbf{Shelf}}} & 6 & \textbf{0.99} $\pm$ 0.01 & \textbf{0.99} $\pm$ 0.01 & \textbf{0.99} $\pm$ 0.01 & \textbf{0.99} $\pm$ 0.01 \\
 & 12 & 0.98 $\pm$ 0.01 & \textbf{0.99} $\pm$ 0.01 & \textbf{0.99} $\pm$ 0.01 & \textbf{0.99} $\pm$ 0.01 \\
 & 18 & 0.95 $\pm$ 0.03 & 0.98 $\pm$ 0.01 & \textbf{0.99} $\pm$ 0.01 & 0.98 $\pm$ 0.01 \\
\bottomrule
\end{tabular}
\caption{SDF guidance with $\alpha=0.5$: success probability ($\uparrow$), with Wilson intervals, for larger step sizes. Bold marks row maxima within this displayed range.}
\label{tab:sweep-sdf050-large}
\end{table}

\Needspace{21\baselineskip}
\subsection{Single-Agent Reduction (RRT-Connect)}
\label{appendix:rrtc}

Table~\ref{tab:rrtc} evaluates whether the multi-robot problem can be solved directly by applying RRT-Connect in the joint configuration space on Dense. Success falls from $0.83$ with $4$ robots to $0.07$ with $10$, while mean runtime rises from $2.97$ to $21.98$ seconds. Time on success conditions on solved instances and therefore does not reflect the cost of unsuccessful searches; mean runtime includes both outcomes. These results expose the practical scaling difficulty of joint-space search under the tested budget, despite its asymptotic completeness properties.

\begin{table}[H]
\centering
\footnotesize
\setlength{\tabcolsep}{3.25pt}
\setlength{\aboverulesep}{0pt}
\setlength{\belowrulesep}{0pt}
\begin{tabular}{c|ccc}
\toprule
Robots & Success $\uparrow$ & Time on success $\downarrow$ & Mean runtime $\downarrow$ \\
\midrule
4  & 0.83 $\pm$ 0.13 & 2.07 $\pm$ 0.49 & 2.97 $\pm$ 0.66 \\
6  & 0.34 $\pm$ 0.17 & 4.55 $\pm$ 1.91 & 11.35 $\pm$ 1.42 \\
8  & 0.14 $\pm$ 0.11 & 3.37 $\pm$ 2.87 & 17.30 $\pm$ 1.10 \\
10 & 0.07 $\pm$ 0.07 & -- & 21.98 $\pm$ 1.09 \\
\bottomrule
\end{tabular}
\caption{RRT-Connect on Dense. Success is a fraction; times are in seconds. Conditional solve time is reported for successful trials when available; ``--'' denotes an unreported value.}
\label{tab:rrtc}
\end{table}

\Needspace{25\baselineskip}
\subsection{Coarse Planner Ablation}
\label{appendix:coarse_planner}

We compare straight-line interpolation, Brownian bridges with maximum variance $\sigma^2=0.025^2$, Voronoi roadmaps, and RRT. The tables report success (S, in \%), runtime (T, seconds), and path length (P). Path length must be interpreted jointly with success, since short trajectories can remain infeasible.

\Needspace{26\baselineskip}
{\bf TFDP with simple initializations.} Table~\ref{tab:coarse-tfdp-simple} tests refinement from straight lines and Brownian bridges. Straight lines achieve $89$--$98\%$ success on Basic, but both initializers reach only $0$--$17\%$ on Room and Shelf. Brownian perturbations do not consistently improve feasibility and reduce Dense success with $18$ robots from $68\%$ to $45\%$. Thus, local trajectory variation alone does not resolve the routing difficulty of structured maps; the short runtimes and paths in failed cases should not be interpreted as better solutions.

\begin{table}[H]
\centering
\footnotesize
\setlength{\tabcolsep}{3.25pt}
\setlength{\aboverulesep}{0pt}
\setlength{\belowrulesep}{0pt}
\begin{tabular}{ll|>{\columncolor{sidgray}}c>{\columncolor{sidgray}}c>{\columncolor{sidgray}}c|>{\columncolor{sidgray}}c>{\columncolor{sidgray}}c>{\columncolor{sidgray}}c}
\toprule
Map & Robots & \multicolumn{3}{>{\columncolor{sidgray}}c|}{Linear} & \multicolumn{3}{>{\columncolor{sidgray}}c}{Brownian bridge} \\
& & S$\uparrow$ & T$\downarrow$ & P$\downarrow$ & S$\uparrow$ & T$\downarrow$ & P$\downarrow$ \\
\midrule
\multirow{3}{*}{\rotatebox[origin=c]{90}{\textbf{Basic}}} & 6 & \textbf{98} & 0.15 $\pm$ 0.00 & \textbf{1.01} $\pm$ 0.00 & 91 & \textbf{0.14} $\pm$ 0.01 & 1.02 $\pm$ 0.00 \\
 & 12 & \textbf{96} & 0.19 $\pm$ 0.00 & \textbf{1.02} $\pm$ 0.00 & 85 & \textbf{0.17} $\pm$ 0.01 & \textbf{1.02} $\pm$ 0.00 \\
 & 18 & \textbf{89} & 0.25 $\pm$ 0.00 & \textbf{1.02} $\pm$ 0.00 & 80 & \textbf{0.23} $\pm$ 0.01 & 1.03 $\pm$ 0.00 \\
\hline
\multirow{3}{*}{\rotatebox[origin=c]{90}{\textbf{Dense}}} & 6 & \textbf{92} & 0.16 $\pm$ 0.00 & \textbf{1.05} $\pm$ 0.00 & 86 & \textbf{0.15} $\pm$ 0.01 & 1.07 $\pm$ 0.00 \\
 & 12 & \textbf{87} & 0.23 $\pm$ 0.00 & \textbf{1.06} $\pm$ 0.00 & 72 & \textbf{0.21} $\pm$ 0.01 & 1.07 $\pm$ 0.00 \\
 & 18 & \textbf{68} & 0.33 $\pm$ 0.00 & \textbf{1.07} $\pm$ 0.00 & 45 & \textbf{0.30} $\pm$ 0.01 & 1.09 $\pm$ 0.00 \\
\hline
\multirow{3}{*}{\rotatebox[origin=c]{90}{\textbf{Room}}} & 6 & \textbf{7} & 0.12 $\pm$ 0.00 & \textbf{1.08} $\pm$ 0.00 & \textbf{7} & \textbf{0.11} $\pm$ 0.00 & 1.10 $\pm$ 0.00 \\
 & 12 & \textbf{0} & 0.15 $\pm$ 0.00 & 1.09 $\pm$ 0.00 & \textbf{0} & \textbf{0.14} $\pm$ 0.00 & 1.10 $\pm$ 0.00 \\
 & 18 & \textbf{0} & 0.19 $\pm$ 0.00 & 1.10 $\pm$ 0.00 & \textbf{0} & \textbf{0.18} $\pm$ 0.01 & 1.11 $\pm$ 0.00 \\
\hline
\multirow{3}{*}{\rotatebox[origin=c]{90}{\textbf{Shelf}}} & 6 & 16 & 0.12 $\pm$ 0.00 & \textbf{1.02} $\pm$ 0.00 & \textbf{17} & \textbf{0.11} $\pm$ 0.00 & 1.04 $\pm$ 0.00 \\
 & 12 & \textbf{4} & 0.15 $\pm$ 0.00 & \textbf{1.03} $\pm$ 0.00 & 3 & \textbf{0.14} $\pm$ 0.01 & 1.04 $\pm$ 0.00 \\
 & 18 & \textbf{0} & 0.19 $\pm$ 0.00 & 1.04 $\pm$ 0.00 & \textbf{0} & \textbf{0.17} $\pm$ 0.01 & 1.05 $\pm$ 0.00 \\
\bottomrule
\end{tabular}
\caption{TFDP with linear and Brownian-bridge initialization. Bold compares the two initializers within each row and metric; path length is ranked only for initializers with nonzero success.}
\label{tab:coarse-tfdp-simple}
\end{table}

\Needspace{26\baselineskip}
{\bf TFDP with obstacle-aware initializations.} Table~\ref{tab:coarse-tfdp-roadmap} shows that Voronoi and RRT restore $99$--$100\%$ success on Room and Shelf. Voronoi yields lower total runtime in every row, while RRT yields shorter paths on the structured maps: on Room with $18$ robots, path length decreases from $2.00$ to $1.56$, at a runtime increase from $0.82$ to $2.06$ seconds. On Dense with $18$ robots, RRT also raises success from $77\%$ to $92\%$. Initialization therefore affects feasibility, efficiency, and path quality.

\begin{table}[H]
\centering
\footnotesize
\setlength{\tabcolsep}{3.25pt}
\setlength{\aboverulesep}{0pt}
\setlength{\belowrulesep}{0pt}
\begin{tabular}{ll|>{\columncolor{sidgray}}c>{\columncolor{sidgray}}c>{\columncolor{sidgray}}c|>{\columncolor{sidgray}}c>{\columncolor{sidgray}}c>{\columncolor{sidgray}}c}
\toprule
Map & Robots & \multicolumn{3}{>{\columncolor{sidgray}}c|}{Voronoi} & \multicolumn{3}{>{\columncolor{sidgray}}c}{RRT} \\
& & S$\uparrow$ & T$\downarrow$ & P$\downarrow$ & S$\uparrow$ & T$\downarrow$ & P$\downarrow$ \\
\midrule
\multirow{3}{*}{\rotatebox[origin=c]{90}{\textbf{Basic}}} & 6 & \textbf{100} & \textbf{0.30} $\pm$ 0.00 & 1.20 $\pm$ 0.00 & \textbf{100} & 0.40 $\pm$ 0.02 & \textbf{1.09} $\pm$ 0.01 \\
 & 12 & \textbf{100} & \textbf{0.47} $\pm$ 0.00 & 1.22 $\pm$ 0.00 & 99 & 0.54 $\pm$ 0.02 & \textbf{1.09} $\pm$ 0.00 \\
 & 18 & 99 & \textbf{0.71} $\pm$ 0.00 & 1.24 $\pm$ 0.00 & \textbf{100} & 0.82 $\pm$ 0.04 & \textbf{1.12} $\pm$ 0.00 \\
\hline
\multirow{3}{*}{\rotatebox[origin=c]{90}{\textbf{Dense}}} & 6 & 96 & \textbf{0.39} $\pm$ 0.00 & \textbf{1.29} $\pm$ 0.01 & \textbf{99} & 0.67 $\pm$ 0.02 & 1.32 $\pm$ 0.02 \\
 & 12 & 90 & \textbf{0.63} $\pm$ 0.00 & \textbf{1.32} $\pm$ 0.00 & \textbf{98} & 1.13 $\pm$ 0.02 & 1.37 $\pm$ 0.01 \\
 & 18 & 77 & \textbf{0.90} $\pm$ 0.00 & \textbf{1.34} $\pm$ 0.00 & \textbf{92} & 1.91 $\pm$ 0.04 & 1.39 $\pm$ 0.01 \\
\hline
\multirow{3}{*}{\rotatebox[origin=c]{90}{\textbf{Room}}} & 6 & \textbf{100} & \textbf{0.31} $\pm$ 0.00 & 1.93 $\pm$ 0.01 & \textbf{100} & 0.71 $\pm$ 0.01 & \textbf{1.43} $\pm$ 0.01 \\
 & 12 & \textbf{100} & \textbf{0.52} $\pm$ 0.00 & 1.96 $\pm$ 0.01 & \textbf{100} & 1.27 $\pm$ 0.02 & \textbf{1.53} $\pm$ 0.01 \\
 & 18 & \textbf{100} & \textbf{0.82} $\pm$ 0.00 & 2.00 $\pm$ 0.00 & \textbf{100} & 2.06 $\pm$ 0.03 & \textbf{1.56} $\pm$ 0.01 \\
\hline
\multirow{3}{*}{\rotatebox[origin=c]{90}{\textbf{Shelf}}} & 6 & \textbf{100} & \textbf{0.31} $\pm$ 0.00 & 1.68 $\pm$ 0.01 & \textbf{100} & 0.53 $\pm$ 0.01 & \textbf{1.30} $\pm$ 0.01 \\
 & 12 & \textbf{100} & \textbf{0.51} $\pm$ 0.00 & 1.73 $\pm$ 0.01 & 99 & 0.90 $\pm$ 0.02 & \textbf{1.34} $\pm$ 0.01 \\
 & 18 & \textbf{100} & \textbf{0.79} $\pm$ 0.00 & 1.76 $\pm$ 0.00 & 99 & 1.52 $\pm$ 0.04 & \textbf{1.38} $\pm$ 0.01 \\
\bottomrule
\end{tabular}
\caption{TFDP with Voronoi and RRT initialization. Bold compares the two initializers within each row and metric; path length is ranked only for initializers with nonzero success.}
\label{tab:coarse-tfdp-roadmap}
\end{table}

\Needspace{26\baselineskip}
{\bf MPPI with simple initializations.} Table~\ref{tab:coarse-mppi-simple} asks whether the same initialization difficulty appears with sampling-based updates. MPPI reaches $96$--$98\%$ success on Basic with straight lines, but at most $17\%$ on Room and $25\%$ on Shelf. With $18$ robots on Dense, Brownian bridges reduce success from $59\%$ to $42\%$. The similar pattern for MPPI and TFDP indicates that poor initial routes are a shared obstacle for local refinement, rather than a limitation specific to analytical scores.

\begin{table}[H]
\centering
\footnotesize
\setlength{\tabcolsep}{3.25pt}
\setlength{\aboverulesep}{0pt}
\setlength{\belowrulesep}{0pt}
\begin{tabular}{ll|ccc|ccc}
\toprule
Map & Robots & \multicolumn{3}{c|}{Linear} & \multicolumn{3}{c}{Brownian bridge} \\
& & S$\uparrow$ & T$\downarrow$ & P$\downarrow$ & S$\uparrow$ & T$\downarrow$ & P$\downarrow$ \\
\midrule
\multirow{3}{*}{\rotatebox[origin=c]{90}{\textbf{Basic}}} & 6 & 98 & 0.65 $\pm$ 0.01 & \textbf{1.03} $\pm$ 0.00 & \textbf{99} & \textbf{0.60} $\pm$ 0.03 & 1.04 $\pm$ 0.00 \\
 & 12 & \textbf{97} & 0.65 $\pm$ 0.01 & \textbf{1.04} $\pm$ 0.00 & 95 & \textbf{0.61} $\pm$ 0.03 & \textbf{1.04} $\pm$ 0.00 \\
 & 18 & \textbf{96} & 0.68 $\pm$ 0.01 & \textbf{1.05} $\pm$ 0.00 & 87 & \textbf{0.61} $\pm$ 0.03 & \textbf{1.05} $\pm$ 0.00 \\
\hline
\multirow{3}{*}{\rotatebox[origin=c]{90}{\textbf{Dense}}} & 6 & \textbf{92} & 0.66 $\pm$ 0.01 & \textbf{1.08} $\pm$ 0.01 & 83 & \textbf{0.62} $\pm$ 0.03 & 1.10 $\pm$ 0.00 \\
 & 12 & \textbf{78} & 0.67 $\pm$ 0.02 & \textbf{1.12} $\pm$ 0.00 & 63 & \textbf{0.62} $\pm$ 0.04 & 1.14 $\pm$ 0.00 \\
 & 18 & \textbf{59} & 0.70 $\pm$ 0.01 & \textbf{1.14} $\pm$ 0.00 & 42 & \textbf{0.64} $\pm$ 0.04 & 1.15 $\pm$ 0.00 \\
\hline
\multirow{3}{*}{\rotatebox[origin=c]{90}{\textbf{Room}}} & 6 & \textbf{17} & 0.65 $\pm$ 0.01 & \textbf{1.13} $\pm$ 0.00 & 15 & \textbf{0.60} $\pm$ 0.03 & \textbf{1.13} $\pm$ 0.00 \\
 & 12 & \textbf{2} & 0.66 $\pm$ 0.01 & \textbf{1.14} $\pm$ 0.00 & 0 & \textbf{0.61} $\pm$ 0.03 & 1.14 $\pm$ 0.00 \\
 & 18 & \textbf{0} & 0.66 $\pm$ 0.01 & 1.15 $\pm$ 0.00 & \textbf{0} & \textbf{0.61} $\pm$ 0.03 & 1.15 $\pm$ 0.00 \\
\hline
\multirow{3}{*}{\rotatebox[origin=c]{90}{\textbf{Shelf}}} & 6 & \textbf{25} & 0.62 $\pm$ 0.01 & 1.06 $\pm$ 0.00 & 23 & \textbf{0.59} $\pm$ 0.03 & \textbf{1.05} $\pm$ 0.00 \\
 & 12 & \textbf{6} & 0.63 $\pm$ 0.01 & \textbf{1.07} $\pm$ 0.00 & 2 & \textbf{0.60} $\pm$ 0.03 & 1.08 $\pm$ 0.00 \\
 & 18 & \textbf{1} & 0.62 $\pm$ 0.02 & \textbf{1.08} $\pm$ 0.00 & 0 & \textbf{0.59} $\pm$ 0.03 & 1.09 $\pm$ 0.00 \\
\bottomrule
\end{tabular}
\caption{MPPI with linear and Brownian-bridge initialization. Bold compares the two initializers within each row and metric; path length is ranked only for initializers with nonzero success.}
\label{tab:coarse-mppi-simple}
\end{table}

\Needspace{26\baselineskip}
{\bf MPPI with obstacle-aware initializations.} Table~\ref{tab:coarse-mppi-roadmap} reports $98$--$100\%$ success on Room and Shelf, confirming that informative initial routes also benefit MPPI. RRT shortens paths on both layouts, generally at higher runtime; on Room with $18$ robots, path length decreases from $2.05$ to $1.77$ while runtime increases from $1.09$ to $1.66$ seconds. On Dense with $18$ robots, RRT instead improves success from $77\%$ to $94\%$ at the cost of longer paths. No initializer dominates all three metrics.

\begin{table}[H]
\centering
\footnotesize
\setlength{\tabcolsep}{3.25pt}
\setlength{\aboverulesep}{0pt}
\setlength{\belowrulesep}{0pt}
\begin{tabular}{ll|ccc|ccc}
\toprule
Map & Robots & \multicolumn{3}{c|}{Voronoi} & \multicolumn{3}{c}{RRT} \\
& & S$\uparrow$ & T$\downarrow$ & P$\downarrow$ & S$\uparrow$ & T$\downarrow$ & P$\downarrow$ \\
\midrule
\multirow{3}{*}{\rotatebox[origin=c]{90}{\textbf{Basic}}} & 6 & \textbf{100} & 0.69 $\pm$ 0.03 & 1.21 $\pm$ 0.01 & \textbf{100} & \textbf{0.63} $\pm$ 0.01 & \textbf{1.13} $\pm$ 0.01 \\
 & 12 & \textbf{100} & 0.80 $\pm$ 0.03 & 1.26 $\pm$ 0.00 & \textbf{100} & \textbf{0.72} $\pm$ 0.01 & \textbf{1.16} $\pm$ 0.01 \\
 & 18 & \textbf{99} & 0.93 $\pm$ 0.03 & 1.29 $\pm$ 0.00 & \textbf{99} & \textbf{0.81} $\pm$ 0.02 & \textbf{1.19} $\pm$ 0.00 \\
\hline
\multirow{3}{*}{\rotatebox[origin=c]{90}{\textbf{Dense}}} & 6 & 95 & \textbf{0.79} $\pm$ 0.03 & \textbf{1.34} $\pm$ 0.01 & \textbf{99} & 0.85 $\pm$ 0.02 & 1.46 $\pm$ 0.02 \\
 & 12 & 89 & \textbf{0.95} $\pm$ 0.03 & \textbf{1.40} $\pm$ 0.00 & \textbf{99} & 1.09 $\pm$ 0.02 & 1.53 $\pm$ 0.01 \\
 & 18 & 77 & \textbf{1.14} $\pm$ 0.04 & \textbf{1.44} $\pm$ 0.00 & \textbf{94} & 1.40 $\pm$ 0.02 & 1.59 $\pm$ 0.01 \\
\hline
\multirow{3}{*}{\rotatebox[origin=c]{90}{\textbf{Room}}} & 6 & \textbf{100} & \textbf{0.74} $\pm$ 0.03 & 1.90 $\pm$ 0.01 & \textbf{100} & 0.93 $\pm$ 0.01 & \textbf{1.57} $\pm$ 0.01 \\
 & 12 & \textbf{100} & \textbf{0.90} $\pm$ 0.03 & 1.98 $\pm$ 0.00 & \textbf{100} & 1.31 $\pm$ 0.02 & \textbf{1.69} $\pm$ 0.02 \\
 & 18 & \textbf{100} & \textbf{1.09} $\pm$ 0.03 & 2.05 $\pm$ 0.00 & 99 & 1.66 $\pm$ 0.02 & \textbf{1.77} $\pm$ 0.01 \\
\hline
\multirow{3}{*}{\rotatebox[origin=c]{90}{\textbf{Shelf}}} & 6 & \textbf{100} & \textbf{0.74} $\pm$ 0.03 & 1.64 $\pm$ 0.01 & \textbf{100} & \textbf{0.74} $\pm$ 0.02 & \textbf{1.46} $\pm$ 0.02 \\
 & 12 & \textbf{100} & \textbf{0.91} $\pm$ 0.03 & 1.73 $\pm$ 0.01 & \textbf{100} & 1.00 $\pm$ 0.02 & \textbf{1.54} $\pm$ 0.02 \\
 & 18 & \textbf{100} & \textbf{1.10} $\pm$ 0.03 & 1.80 $\pm$ 0.00 & 98 & 1.24 $\pm$ 0.02 & \textbf{1.62} $\pm$ 0.01 \\
\bottomrule
\end{tabular}
\caption{MPPI with Voronoi and RRT initialization. Bold compares the two initializers within each row and metric; path length is ranked only for initializers with nonzero success.}
\label{tab:coarse-mppi-roadmap}
\end{table}

\Needspace{26\baselineskip}
\subsection{MPPI CPU Runtime}
\label{appendix:mppi_cpu}

Table~\ref{tab:mppi_cpu_runtime} compares MPPI runtime on the CPU used for TFDP. As robot count grows from $6$ to $18$, runtime increases from $8.18$ to $32.36$ seconds on Basic and from $10.84$ to $40.98$ seconds on Dense. All entries exceed the corresponding TFDP times in Table~\ref{tab:results}. Thus, TFDP retains its runtime advantage on shared hardware, although this comparison does not isolate the cost of individual algorithmic components.

\begin{table}[H]
\centering
\footnotesize
\setlength{\tabcolsep}{3.25pt}
\setlength{\aboverulesep}{0pt}
\setlength{\belowrulesep}{0pt}
\begin{tabular}{ll|c}
\toprule
Map & Robots & MPPI time (s) $\downarrow$ \\
\midrule
\multirow{3}{*}{\rotatebox[origin=c]{90}{\textbf{Basic}}}
& 6  & 8.18 $\pm$ 0.13 \\
& 12 & 19.43 $\pm$ 0.26 \\
& 18 & 32.36 $\pm$ 0.47 \\
\hline
\multirow{3}{*}{\rotatebox[origin=c]{90}{\textbf{Dense}}}
& 6  & 10.84 $\pm$ 0.26 \\
& 12 & 24.85 $\pm$ 0.54 \\
& 18 & 40.98 $\pm$ 0.86 \\
\hline
\multirow{3}{*}{\rotatebox[origin=c]{90}{\textbf{Room}}}
& 6  & 9.87 $\pm$ 0.20 \\
& 12 & 22.88 $\pm$ 0.41 \\
& 18 & 38.62 $\pm$ 0.63 \\
\hline
\multirow{3}{*}{\rotatebox[origin=c]{90}{\textbf{Shelf}}}
& 6  & 9.12 $\pm$ 0.17 \\
& 12 & 21.39 $\pm$ 0.35 \\
& 18 & 36.31 $\pm$ 0.53 \\
\bottomrule
\end{tabular}
\caption{MPPI runtime on the same CPU as TFDP; lower is better.}
\label{tab:mppi_cpu_runtime}
\end{table}

\Needspace{34\baselineskip}
\subsection{Effect of Noise}
\label{appendix:effect_of_noise}

Table~\ref{tab:effect-noise} compares stochastic refinement with deterministic annealed updates. Removing noise matches or slightly improves success on the standard maps. In the crowded stress tests, noise instead improves the point estimates: with $12$ robots, Bottleneck rises from $0.38$ to $0.45$, and Standoff from $0.39$ to $0.45$. This suggests that stochastic exploration is most useful when many robots compete for limited passage space. The uncertainty ranges overlap, so these gains describe observed trends without establishing statistical significance.

\begin{table}[H]
\centering
\footnotesize
\setlength{\tabcolsep}{3.25pt}
\setlength{\aboverulesep}{0pt}
\setlength{\belowrulesep}{0pt}
\begin{tabular}{l|>{\columncolor{sidgray}}c>{\columncolor{sidgray}}c}
\toprule
Scenario & TFDP, noisy $\uparrow$ & TFDP, noiseless $\uparrow$ \\
\midrule
Dense (6 agents) & \textbf{0.99} $\pm$ 0.01 & \textbf{0.99} $\pm$ 0.01 \\
Dense (12 agents) & \textbf{0.98} $\pm$ 0.02 & \textbf{0.98} $\pm$ 0.01 \\
Dense (18 agents) & 0.93 $\pm$ 0.03 & \textbf{0.95} $\pm$ 0.02 \\
\hline
Basic (6 agents) & 0.99 $\pm$ 0.01 & \textbf{1.00} $\pm$ 0.00 \\
Basic (12 agents) & \textbf{0.99} $\pm$ 0.01 & \textbf{0.99} $\pm$ 0.01 \\
Basic (18 agents) & 0.99 $\pm$ 0.01 & \textbf{1.00} $\pm$ 0.00 \\
\hline
Room (6 agents) & 0.99 $\pm$ 0.01 & \textbf{1.00} $\pm$ 0.00 \\
Room (12 agents) & \textbf{0.99} $\pm$ 0.01 & \textbf{0.99} $\pm$ 0.01 \\
Room (18 agents) & \textbf{0.99} $\pm$ 0.01 & \textbf{0.99} $\pm$ 0.01 \\
\hline
Shelf (6 agents) & 0.99 $\pm$ 0.01 & \textbf{1.00} $\pm$ 0.00 \\
Shelf (12 agents) & \textbf{0.99} $\pm$ 0.01 & \textbf{0.99} $\pm$ 0.01 \\
Shelf (18 agents) & 0.98 $\pm$ 0.01 & \textbf{0.99} $\pm$ 0.01 \\
\hline
Bottleneck (2 agents) & \textbf{0.99} $\pm$ 0.01 & 0.98 $\pm$ 0.01 \\
Bottleneck (4 agents) & 0.94 $\pm$ 0.03 & \textbf{0.95} $\pm$ 0.03 \\
Bottleneck (6 agents) & \textbf{0.90} $\pm$ 0.04 & \textbf{0.90} $\pm$ 0.04 \\
Bottleneck (8 agents) & \textbf{0.81} $\pm$ 0.05 & 0.75 $\pm$ 0.05 \\
Bottleneck (12 agents) & \textbf{0.45} $\pm$ 0.06 & 0.38 $\pm$ 0.06 \\
\hline
Standoff (2 agents) & \textbf{0.99} $\pm$ 0.01 & \textbf{0.99} $\pm$ 0.01 \\
Standoff (4 agents) & \textbf{0.94} $\pm$ 0.03 & 0.93 $\pm$ 0.03 \\
Standoff (6 agents) & \textbf{0.90} $\pm$ 0.04 & 0.88 $\pm$ 0.04 \\
Standoff (8 agents) & \textbf{0.81} $\pm$ 0.05 & 0.76 $\pm$ 0.05 \\
Standoff (12 agents) & \textbf{0.45} $\pm$ 0.06 & 0.39 $\pm$ 0.06 \\
\bottomrule
\end{tabular}
\caption{Success fractions with noisy and deterministic annealed refinement. Bold marks the higher central value in each row, including ties.}
\label{tab:effect-noise}
\end{table}

\Needspace{29\baselineskip}
\subsection{Effect of \texorpdfstring{$\sigma$ and $T$}{sigma and T}}
\label{appendix:sigma_and_t}

We vary the initial smoothing scale $\sigma$ and temperature $T$, which are exponentially annealed to $\sigma=0.001$ and $T=0.0001$. These experiments use straight-line initialization to examine whether the refinement schedule can compensate for a simple initial trajectory. Rows vary $\sigma$, columns vary $T$, and each entry reports the success fraction.

\Needspace{22\baselineskip}
{\bf Dense maps.} Table~\ref{tab:sigma-temperature-dense} shows that smoothing and temperature must be selected jointly. At $\sigma=0.05$, increasing $T$ from $0.0001$ to $0.8$ raises success from $0.40$ to $0.72$; at $\sigma=0.2$, the same change reduces it from $0.64$ to $0.04$. The maximum of $0.72$ occurs at several parameter pairs. Additional exploration can therefore improve refinement from straight lines, but excessive noise at large smoothing scales can sharply reduce feasibility.

\begin{table}[H]
\centering
\footnotesize
\setlength{\tabcolsep}{3.25pt}
\setlength{\aboverulesep}{0pt}
\setlength{\belowrulesep}{0pt}
\begin{tabular}{c|>{\columncolor{sidgray}}c>{\columncolor{sidgray}}c>{\columncolor{sidgray}}c>{\columncolor{sidgray}}c>{\columncolor{sidgray}}c>{\columncolor{sidgray}}c>{\columncolor{sidgray}}c>{\columncolor{sidgray}}c>{\columncolor{sidgray}}c>{\columncolor{sidgray}}c}
\toprule
$\sigma\backslash T$ & 0.0001 & 0.0005 & 0.001 & 0.005 & 0.01 & 0.05 & 0.1 & 0.3 & 0.5 & 0.8 \\
\midrule
0.01  & 0.28 & 0.28 & 0.24 & 0.28 & 0.24 & 0.28 & 0.28 & 0.28 & 0.28 & \textbf{0.40} \\
0.025 & 0.32 & 0.32 & 0.36 & 0.36 & 0.36 & 0.32 & 0.36 & 0.36 & \textbf{0.56} & \textbf{0.56} \\
0.05  & 0.40 & 0.40 & 0.40 & 0.36 & 0.40 & 0.36 & 0.40 & 0.64 & 0.68 & \textbf{0.72} \\
0.075 & 0.48 & 0.48 & 0.48 & 0.48 & 0.48 & 0.44 & 0.44 & 0.60 & \textbf{0.68} & 0.60 \\
0.1   & 0.44 & 0.44 & 0.44 & 0.44 & 0.44 & 0.44 & 0.44 & \textbf{0.72} & 0.64 & 0.40 \\
0.125 & 0.48 & 0.48 & 0.48 & 0.48 & 0.48 & 0.52 & 0.60 & 0.60 & \textbf{0.64} & 0.32 \\
0.15  & 0.52 & 0.52 & 0.52 & 0.52 & 0.52 & 0.56 & 0.64 & \textbf{0.68} & 0.52 & 0.12 \\
0.175 & 0.60 & 0.64 & 0.64 & 0.56 & 0.56 & 0.64 & \textbf{0.72} & 0.68 & 0.20 & 0.08 \\
0.2   & 0.64 & 0.60 & 0.60 & \textbf{0.68} & 0.64 & 0.60 & 0.64 & 0.52 & 0.12 & 0.04 \\
\bottomrule
\end{tabular}
\caption{TFDP success fraction ($\uparrow$) on Dense. Rows vary initial $\sigma$ and columns vary initial $T$; bold marks the best temperature in each row, including ties.}
\label{tab:sigma-temperature-dense}
\end{table}

\Needspace{22\baselineskip}
{\bf Basic maps.} Table~\ref{tab:sigma-temperature-basic} reaches $0.96$ success at $(\sigma,T)=(0.05,0.8)$ and $(0.1,0.5)$, compared with $0.72$ at the lowest temperature for either smoothing scale. This shows that suitable stochastic refinement can substantially improve a simple initialization in this setting. The benefit is not monotonic: at $\sigma=0.2$, success falls from $0.84$ at $T=0.1$ to $0.08$ at $T=0.8$. Even on Basic, increasing temperature alone is insufficient.

\begin{table}[H]
\centering
\footnotesize
\setlength{\tabcolsep}{3.25pt}
\setlength{\aboverulesep}{0pt}
\setlength{\belowrulesep}{0pt}
\begin{tabular}{c|>{\columncolor{sidgray}}c>{\columncolor{sidgray}}c>{\columncolor{sidgray}}c>{\columncolor{sidgray}}c>{\columncolor{sidgray}}c>{\columncolor{sidgray}}c>{\columncolor{sidgray}}c>{\columncolor{sidgray}}c>{\columncolor{sidgray}}c>{\columncolor{sidgray}}c}
\toprule
$\sigma\backslash T$ & 0.0001 & 0.0005 & 0.001 & 0.005 & 0.01 & 0.05 & 0.1 & 0.3 & 0.5 & 0.8 \\
\midrule
0.01  & 0.72 & 0.72 & 0.72 & 0.72 & 0.72 & 0.72 & 0.72 & 0.72 & \textbf{0.76} & 0.68 \\
0.025 & 0.72 & 0.72 & 0.72 & 0.72 & 0.72 & 0.72 & 0.72 & 0.76 & 0.72 & \textbf{0.88} \\
0.05  & 0.72 & 0.72 & 0.72 & 0.72 & 0.72 & 0.72 & 0.72 & 0.80 & 0.88 & \textbf{0.96} \\
0.075 & 0.72 & 0.72 & 0.72 & 0.72 & 0.72 & 0.76 & 0.76 & 0.84 & \textbf{0.92} & 0.88 \\
0.1   & 0.72 & 0.72 & 0.72 & 0.72 & 0.72 & 0.72 & 0.76 & 0.88 & \textbf{0.96} & 0.76 \\
0.125 & 0.76 & 0.76 & 0.76 & 0.76 & 0.76 & 0.72 & 0.76 & \textbf{0.92} & 0.88 & 0.48 \\
0.15  & 0.76 & 0.76 & 0.76 & 0.76 & 0.76 & 0.76 & \textbf{0.84} & 0.80 & 0.64 & 0.40 \\
0.175 & 0.72 & 0.72 & 0.72 & 0.72 & 0.72 & 0.76 & \textbf{0.92} & 0.76 & 0.64 & 0.20 \\
0.2   & 0.76 & 0.76 & 0.76 & 0.76 & 0.76 & 0.80 & \textbf{0.84} & 0.72 & 0.52 & 0.08 \\
\bottomrule
\end{tabular}
\caption{TFDP success fraction ($\uparrow$) on Basic. Rows vary initial $\sigma$ and columns vary initial $T$; bold marks the best temperature in each row, including ties.}
\label{tab:sigma-temperature-basic}
\end{table}

\Needspace{28\baselineskip}
\subsection{Density/Stress Tests}
\label{appendix:deadlock_prone}

\begin{figure}[H]
    \centering
    \includegraphics[width=0.75\linewidth]{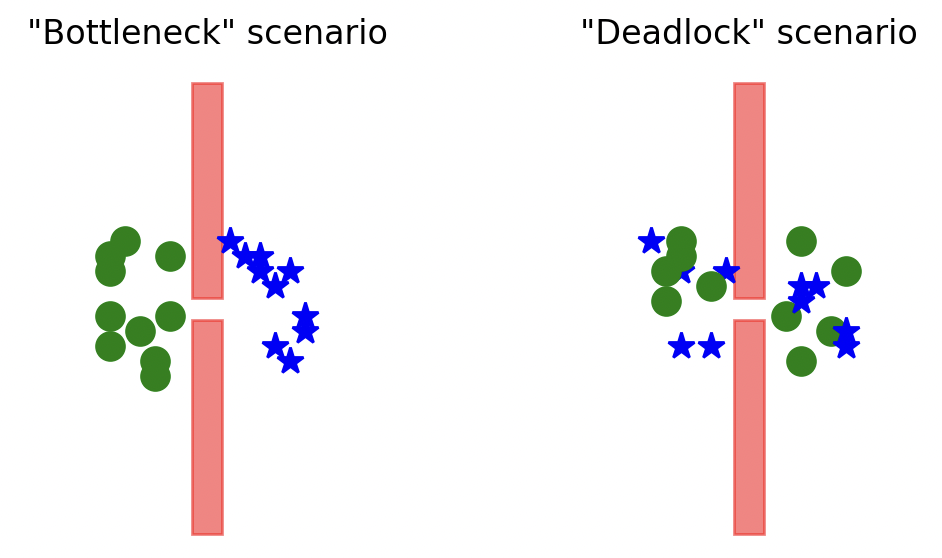}
    \caption{Stress-test layouts: Bottleneck (left) and Standoff (right, labeled ``Deadlock''). Both have an opening wide enough for one robot.}
    \label{fig:stress_test_maps}
\end{figure}

We test maps with randomized start and goal positions and a wall opening wide enough for one robot. In Bottleneck, robots start on the same side; in Standoff, half the start--goal pairs are reversed to create opposing traffic. Figure~\ref{fig:stress_test_maps} shows the layouts.

\Needspace{18\baselineskip}
{\bf Bottleneck.} Table~\ref{tab:stress-bottleneck} measures feasibility when robots must traverse a shared narrow opening in the same direction. TFDP has higher success point estimates at every tested count; with $8$ robots, success is $0.69$ versus MPPI's $0.20$. However, TFDP falls to $0.31$ with $12$ robots, while MPPI reaches $0.02$. The results show a substantial advantage under congestion, but also identify crowded passages as a remaining source of failure.

\begin{table}[H]
\centering
\footnotesize
\setlength{\tabcolsep}{3.25pt}
\setlength{\aboverulesep}{0pt}
\setlength{\belowrulesep}{0pt}
\begin{tabular}{c|c|>{\columncolor{sidgray}}c}
\toprule
Robots & MPPI success $\uparrow$ & TFDP success $\uparrow$ \\
\midrule
2  & 0.96 $\pm$ 0.02 & \textbf{0.98} $\pm$ 0.02 \\
4  & 0.76 $\pm$ 0.05 & \textbf{0.95} $\pm$ 0.03 \\
6  & 0.51 $\pm$ 0.06 & \textbf{0.89} $\pm$ 0.04 \\
8  & 0.20 $\pm$ 0.05 & \textbf{0.69} $\pm$ 0.06 \\
12 & 0.02 $\pm$ 0.02 & \textbf{0.31} $\pm$ 0.06 \\
\bottomrule
\end{tabular}
\caption{Bottleneck success fractions. Bold marks the higher central value in each row.}
\label{tab:stress-bottleneck}
\end{table}

\Needspace{15\baselineskip}
{\bf Standoff.} Table~\ref{tab:stress-standoff} introduces opposing traffic through the same type of opening, requiring robots moving in different directions to coordinate passage. TFDP achieves $0.70$ success with $8$ robots, compared with $0.23$ for MPPI; with $12$ robots, the values fall to $0.32$ and $0.03$. The advantage therefore persists under opposing traffic, while the low success at the largest count shows that analytical refinement does not consistently resolve these tightly coupled interactions.

\begin{table}[H]
\centering
\footnotesize
\setlength{\tabcolsep}{3.25pt}
\setlength{\aboverulesep}{0pt}
\setlength{\belowrulesep}{0pt}
\begin{tabular}{c|c|>{\columncolor{sidgray}}c}
\toprule
Robots & MPPI success $\uparrow$ & TFDP success $\uparrow$ \\
\midrule
2  & 0.96 $\pm$ 0.02 & \textbf{0.99} $\pm$ 0.01 \\
4  & 0.78 $\pm$ 0.05 & \textbf{0.92} $\pm$ 0.03 \\
6  & 0.51 $\pm$ 0.06 & \textbf{0.85} $\pm$ 0.05 \\
8  & 0.23 $\pm$ 0.05 & \textbf{0.70} $\pm$ 0.06 \\
12 & 0.03 $\pm$ 0.02 & \textbf{0.32} $\pm$ 0.06 \\
\bottomrule
\end{tabular}
\caption{Standoff success fractions. Bold marks the higher central value in each row.}
\label{tab:stress-standoff}
\end{table}

\Needspace{30\baselineskip}
\subsection{Locality of Learned Scores}
\label{app:nn_score_dependence_test}

We test how much temporal context is useful for score estimation by training a multilayer perceptron with AdaLN-Zero conditioning on noise level for a simple trajectory distribution. This network is a diagnostic model; TFDP itself remains training-free. Table~\ref{tab:trainable-final-loss} reports final-epoch mean squared error as the temporal window and noise level vary.

{\bf Results.} Expanding the window from $1$ to $3$ reduces error at every noise level, showing that neighboring waypoints carry useful information. At $\sigma=0.05$, error decreases from $0.8679$ to $0.3849$, then reaches its minimum of $0.2988$ at window $7$; window $9$ gives no further improvement. At $\sigma=0.2$, error continues to decrease through window $9$, reaching $0.2037$. Thus, short neighborhoods are effective in this diagnostic, while broader context remains useful at higher noise. This supports the use of local score approximations at low noise, without establishing exact temporal independence or directly measuring downstream planning performance.

\begin{table}[H]
\centering
\footnotesize
\setlength{\tabcolsep}{3.25pt}
\setlength{\aboverulesep}{0pt}
\setlength{\belowrulesep}{0pt}
\begin{tabular}{l|cccc}
\toprule
Window & $\sigma=0.025$ & $\sigma=0.05$ & $\sigma=0.1$ & $\sigma=0.2$ \\
\midrule
1 & 0.8391 & 0.8679 & 0.9317 & 0.9036 \\
3 & 0.4587 & 0.3849 & 0.3572 & 0.3734 \\
5 & \textbf{0.4169} & 0.3118 & 0.2566 & 0.2712 \\
7 & 0.4752 & \textbf{0.2988} & 0.2299 & 0.2334 \\
9 & 0.4464 & 0.3006 & \textbf{0.2184} & \textbf{0.2037} \\
\bottomrule
\end{tabular}
\caption{Final-epoch mean squared error ($\downarrow$) of the diagnostic score estimator. Bold marks the lowest error across window sizes for each noise level.}
\label{tab:trainable-final-loss}
\end{table}

\Needspace{22\baselineskip}
\subsection{Confidence Bounds for 250-Step TFDP Results}
\label{appendix:confidence_intervals}

Table~\ref{tab:tfdp-ci} quantifies uncertainty for TFDP with $250$ denoising steps. Basic and Shelf have $100\%$ observed success at every robot count, but their Wilson lower bounds are $98.5\%$, distinguishing empirical success from guaranteed feasibility. Dense with $18$ robots is the least reliable reported setting, at $96\%$ with a $[92.7,97.9]\%$ interval. Runtime increases with robot count on every map, while mean path length changes more modestly. Together, the estimates support high observed reliability at these scales while making the uncertainty in that conclusion explicit.

\begin{table}[H]
\centering
\footnotesize
\setlength{\tabcolsep}{3.25pt}
\setlength{\aboverulesep}{0pt}
\setlength{\belowrulesep}{0pt}
\begin{tabular}{ll|>{\columncolor{sidgray}}c>{\columncolor{sidgray}}c>{\columncolor{sidgray}}c}
\toprule
Map & Robots & S$\uparrow$ (\%, 95\% CI) & T$\downarrow$ (s) & P$\downarrow$ \\
\midrule
\multirow{3}{*}{\rotatebox[origin=c]{90}{\textbf{Basic}}}
 & 6  & $100\;[\,98.5,\,100.0\,]$ & $0.18 \pm 0.02$ & $1.07 \pm 0.05$ \\
 & 12 & $100\;[\,98.5,\,100.0\,]$ & $0.29 \pm 0.08$ & $1.08 \pm 0.04$ \\
 & 18 & $100\;[\,98.5,\,100.0\,]$ & $0.45 \pm 0.07$ & $1.08 \pm 0.03$ \\
\hline
\multirow{3}{*}{\rotatebox[origin=c]{90}{\textbf{Dense}}}
 & 6  & $100\;[\,98.5,\,100.0\,]$ & $0.35 \pm 0.09$ & $1.26 \pm 0.15$ \\
 & 12 & $\phantom{0}99\;[\,96.8,\,99.7\,]$  & $0.66 \pm 0.07$ & $1.29 \pm 0.09$ \\
 & 18 & $\phantom{0}96\;[\,92.7,\,97.9\,]$  & $1.06 \pm 0.19$ & $1.32 \pm 0.03$ \\
\hline
\multirow{3}{*}{\rotatebox[origin=c]{90}{\textbf{Room}}}
 & 6  & $100\;[\,98.5,\,100.0\,]$ & $0.43 \pm 0.09$ & $1.41 \pm 0.17$ \\
 & 12 & $\phantom{0}99\;[\,96.8,\,99.7\,]$  & $0.78 \pm 0.12$ & $1.45 \pm 0.08$ \\
 & 18 & $\phantom{0}98\;[\,95.4,\,99.2\,]$  & $1.24 \pm 0.23$ & $1.45 \pm 0.07$ \\
\hline
\multirow{3}{*}{\rotatebox[origin=c]{90}{\textbf{Shelf}}}
 & 6  & $100\;[\,98.5,\,100.0\,]$ & $0.27 \pm 0.07$ & $1.23 \pm 0.16$ \\
 & 12 & $100\;[\,98.5,\,100.0\,]$ & $0.51 \pm 0.04$ & $1.25 \pm 0.08$ \\
 & 18 & $100\;[\,98.5,\,100.0\,]$ & $0.84 \pm 0.08$ & $1.26 \pm 0.04$ \\
\bottomrule
\end{tabular}
\caption{TFDP success with $95\%$ Wilson intervals, runtime, and path length. Runtime and path length use mean $\pm 2$ standard errors; gray identifies TFDP results.}
\label{tab:tfdp-ci}
\end{table}

{\bf Confidence intervals.} Success uses $95\%$ Wilson score intervals, which remain within $[0,100]\%$ even near the boundaries. Runtime and path length use mean $\pm 2$ standard errors. These summaries quantify uncertainty in the reported estimates; they do not separate variability due to sampled instances from variability due to random seeds.

\newpage

\clearpage
\subsection{Additional Planning Visualizations}
\label{appendix:planning_visualizations}

\begin{figure}[htbp]

    \centering
  \begin{subfigure}[t]{0.24\linewidth}
    \centering
    \includegraphics[width=\linewidth]{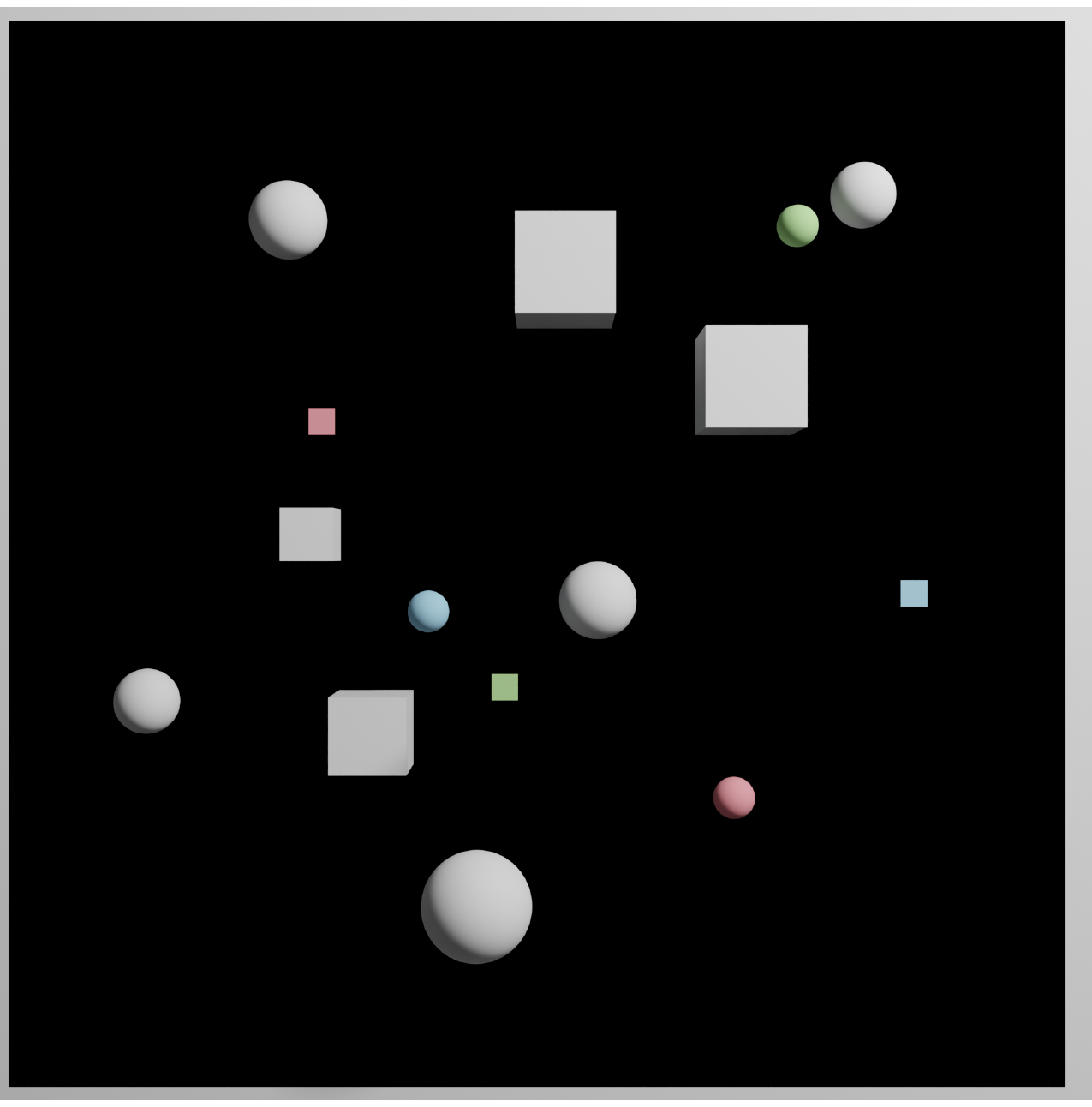}
    
    \caption{Basic.}
    \label{fig:Basic Maps}
  \end{subfigure}
  \hfill
  \begin{subfigure}[t]{0.24\linewidth}
    \centering
    \includegraphics[width=\linewidth]{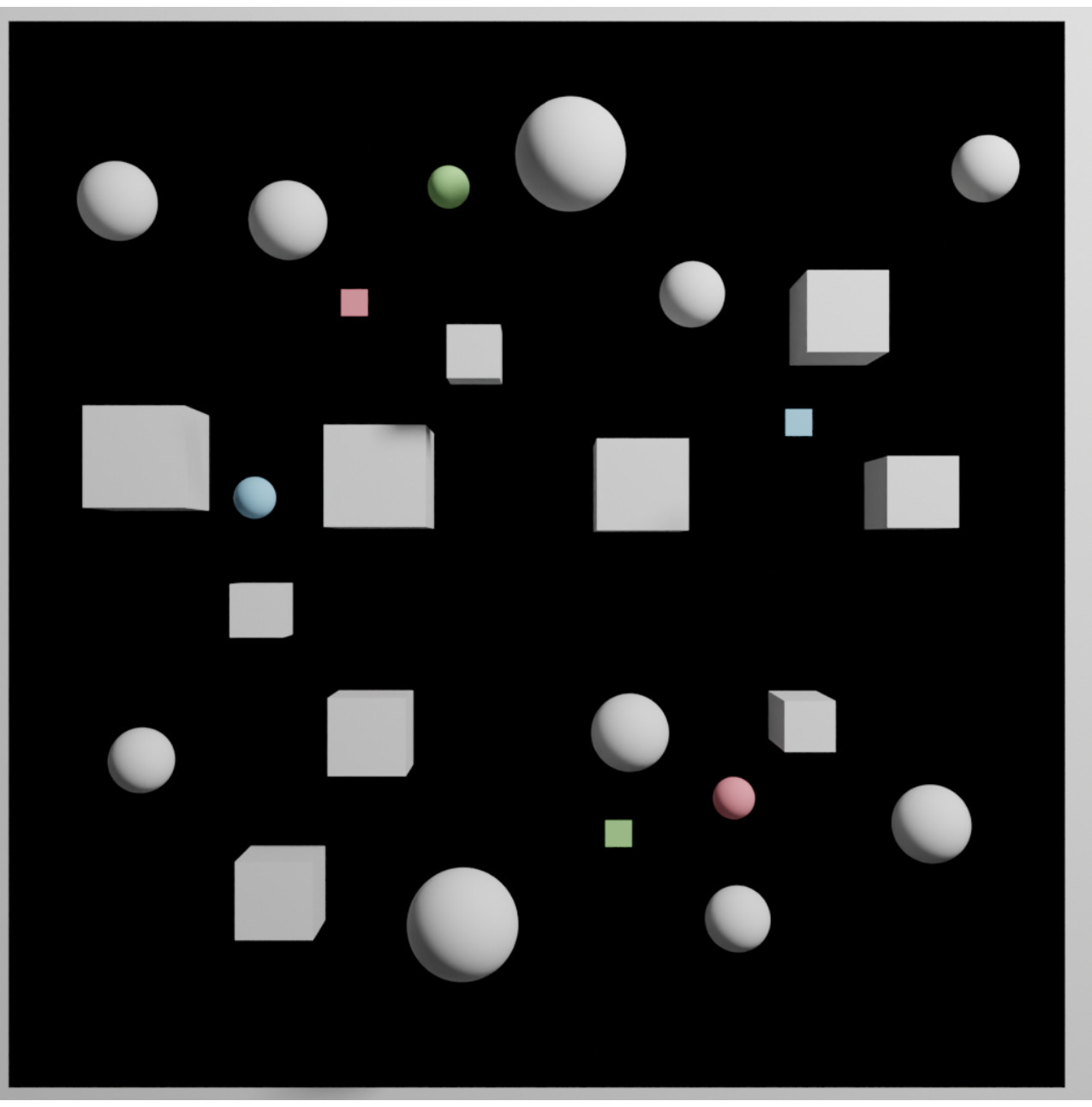}
    
    \caption{Dense.}
    \label{fig:Dense Maps}
  \end{subfigure}
  \hfill
  \begin{subfigure}[t]{0.24\linewidth}
    \centering
    \includegraphics[width=\linewidth]{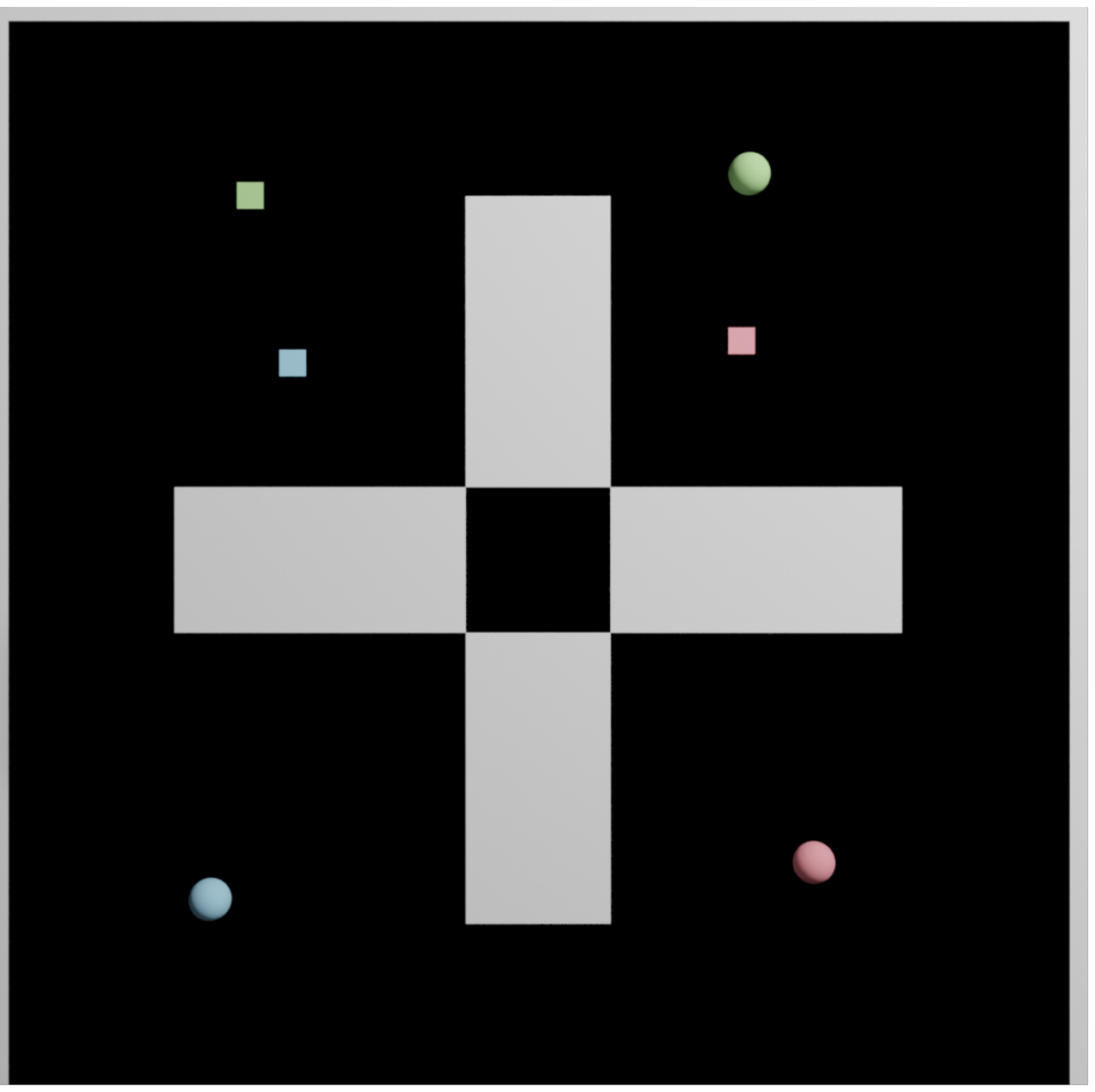}
    
    \caption{Room.}
    \label{fig:Room Maps}
  \end{subfigure}
  \hfill
  \begin{subfigure}[t]{0.24\linewidth}
    \centering
    \includegraphics[width=\linewidth]{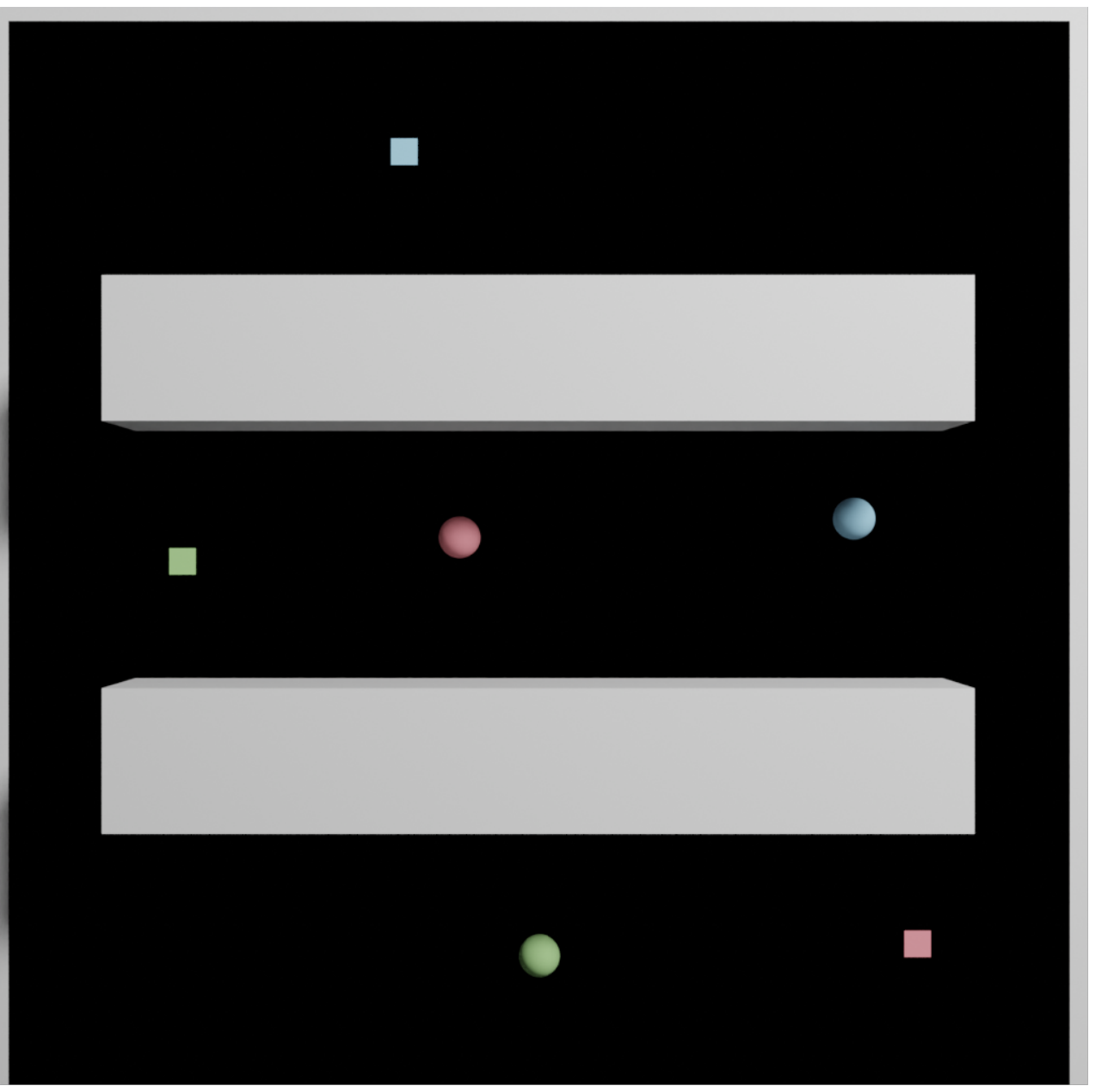}
    
    \caption{Shelf.}
    \label{fig:Shelf Maps}
  \end{subfigure}
    \caption{Example benchmarks used for our experiments.}
    \label{fig:Examples of benchmark}
\end{figure}

{\bf Benchmark geometry.} Figure~\ref{fig:Examples of benchmark} shows the four map types without trajectory overlays. Basic and Dense vary obstacle density, while Room and Shelf impose structured routes through openings and corridors. These layouts probe complementary difficulties: local obstacle avoidance, coordination in crowded spaces, and selection of routes around extended barriers. This distinction helps explain why the simple initializations in Appendix~\ref{appendix:coarse_planner} perform much worse on the structured maps.

\begin{figure}[htbp]
    \centering
    \includegraphics[width=0.65\linewidth,trim={10 10 10 10},clip]{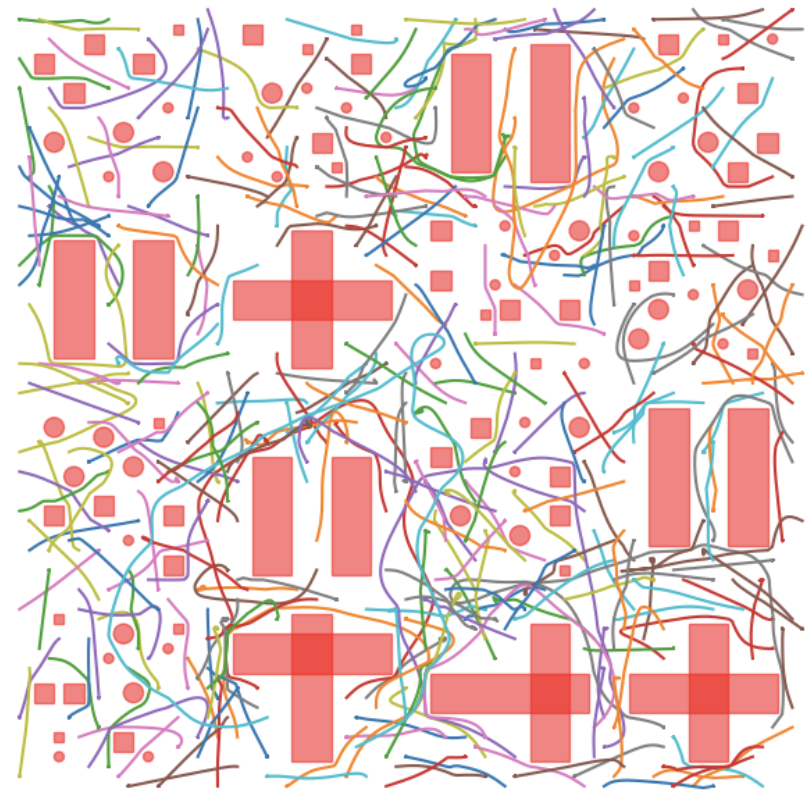}
    \caption{Generated trajectories on the larger map used for the scalability experiment. Colored curves show robot paths; red regions show obstacles.}
    \label{fig:bigmap_plan}
\end{figure}

{\bf Larger environment.} Figure~\ref{fig:bigmap_plan} illustrates trajectories in the large map used for the scalability experiment. The routes span open regions, dense obstacle clusters, and narrow passages, showing the geometric variety handled within one planning instance. The figure complements the scalability experiment with up to $300$ robots, which reports an average runtime of $5.39$ seconds on an A100 GPU. Spatial crossings in this static view do not by themselves indicate collisions, which depend on when robots traverse them.

\clearpage
\subsection{Optimization Timelines}
\label{appendix:optimization_timelines}

\begin{figure}[htbp]
    \centering
    \includegraphics[width=\linewidth]{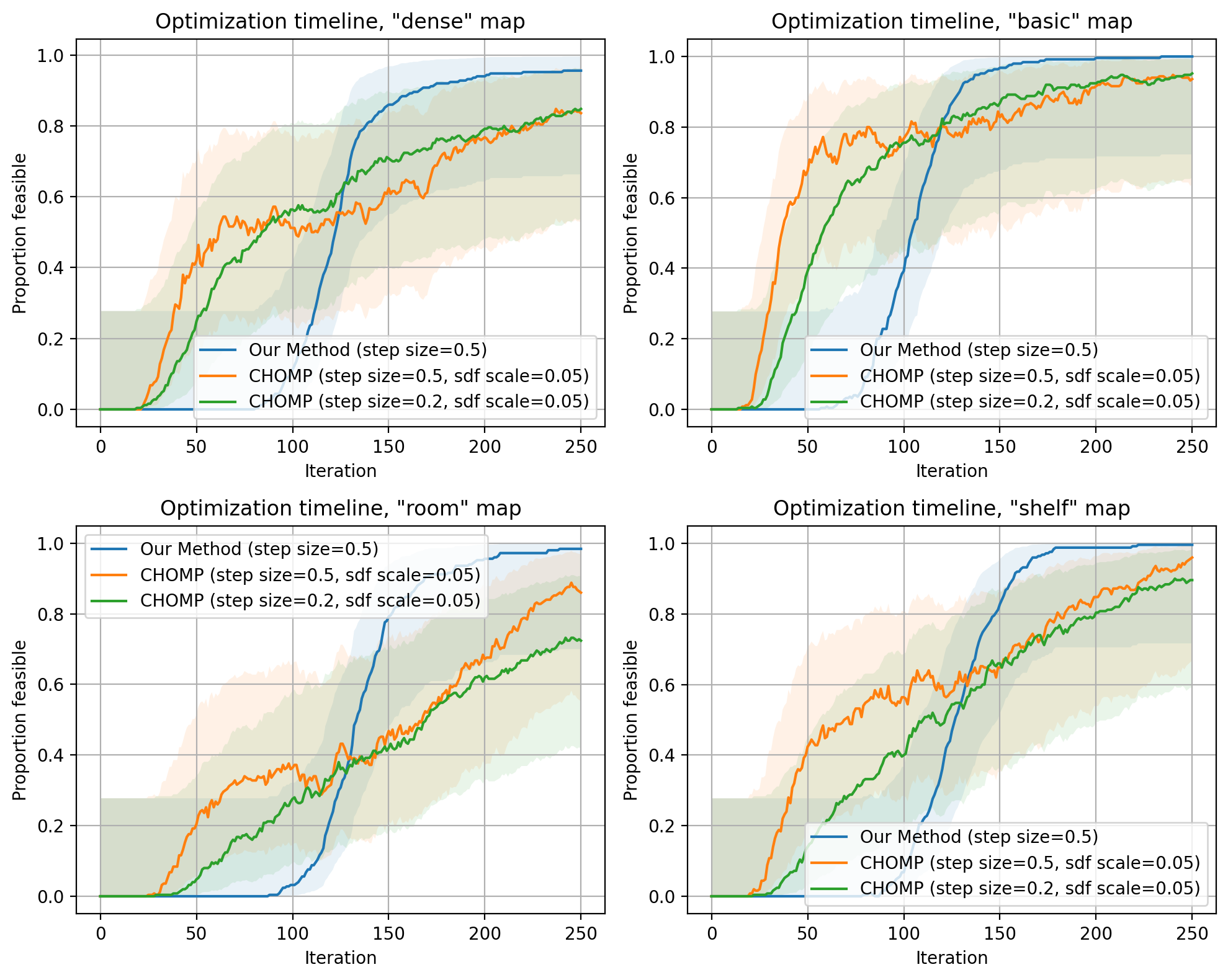}
    \caption{Fraction of feasible plans over refinement iterations on Dense (top left), Basic (top right), Room (bottom left), and Shelf (bottom right). Blue denotes TFDP; orange and green denote CHOMP-style SDF guidance at step sizes $0.5$ and $0.2$, respectively.}
    \label{fig:optimization_timeline}
\end{figure}

Figure~\ref{fig:optimization_timeline} tracks the fraction of feasible plans over $250$ refinement iterations on each benchmark map. TFDP uses $\eta=0.5$; the two CHOMP-style SDF variants use $\alpha=0.05$ with $\eta\in\{0.2,0.5\}$. SDF guidance produces feasible plans earlier, while TFDP improves sharply during the middle of the schedule and finishes with the highest feasible fraction on all four maps. On Dense, the final fractions are approximately $0.96$ for TFDP and $0.84$ for both SDF variants; on Room, they are approximately $0.98$, $0.86$, and $0.73$ for TFDP and SDF with $\eta=0.5$ and $0.2$, respectively. These curves show why final feasibility and early progress need to be assessed separately. Since the horizontal axis counts iterations, the figure describes refinement behavior rather than wall-clock efficiency.

\end{document}